%% file: root.tex
\documentclass[letterpaper, 10 pt, conference]{ieeeconf}  %

\newif\ifanonymous

\anonymousfalse

\IEEEoverridecommandlockouts                              %

\usepackage{cite}
\usepackage{graphicx}
\usepackage{placeins}
\makeatletter
\def\endfigure{\end@float}
\def\endtable{\end@float}
\makeatother
\usepackage[caption=false,font=footnotesize,justification=centering]{subfig}
\usepackage{booktabs}
\usepackage{array}
\usepackage[table]{xcolor}
\usepackage{url}
\usepackage[hidelinks]{hyperref}

\newcommand{\method}{\textsc{ELEVATE}}

\title{\LARGE \bf
ELEVATE: Agentic Controller Acquisition from Task Failure for Quadrupedal Manipulation
}

\ifanonymous
\author{Anonymous Authors}
\else
\author{Shiqi Liu$^{1,2}$, Radu Corcodel$^{2,*}$, and Ding Zhao$^{1}$%
\thanks{This work was fully supported and conducted at Mitsubishi Electric Research Labs (MERL).}%
\thanks{$^{1}$Department of Mechanical Engineering, Carnegie Mellon University,
        Pittsburgh, PA-15213 USA,
        {\tt\small \{shiqiliu, dingzhao\}@andrew.cmu.edu}}%
\thanks{$^{2}$Mitsubishi Electric Research Labs (MERL), Cambridge, MA-02139 USA,
        {\tt\small corcodel@merl.com}}%
\thanks{$^{*}$Project lead. Authors are listed in order of contribution.}%
}
\fi

\begin{document}

\maketitle
\thispagestyle{empty}
\pagestyle{empty}

\input{sessions/abstract.tex}
\input{sessions/intro.tex}
\input{sessions/related_work.tex}

\input{sessions/methodology.tex}

\input{sessions/experiments.tex}
\input{sessions/conclusion.tex}
\input{sessions/references.tex}

\clearpage
\input{sessions/appendix.tex}

\end{document}

%% file: sessions/abstract.tex
\begin{abstract}
Quadrupedal manipulators can extend their workspace through whole-body interaction, but task-planning agents' capabilities remain limited by their available controllers. We present \method{}, a framework that uses persistent failures during task-program refinement to identify and acquire missing whole-body control capabilities. When these failures suggest a missing physical capability, a coding agent formulates and iteratively refines a reinforcement-learning problem, specifying behavior-specific rewards, curricula, and termination conditions. Validated controllers are added to the capability library and composed with existing controllers to retry the task. Starting from a single pretrained loco-manipulation controller, \method{} autonomously acquires controllers for wall-supported elevation, supported manipulation, and wall descent across a sequence of seven simulation tasks. These controllers are reused in subsequent reaching and object-manipulation tasks. On the five tasks beyond the initial controller's workspace, \method{} achieves mean success rates of 98.4\%--100\% across five independent runs, while coding-agent baselines with a fixed controller library fail to complete these tasks. The synthesized task programs, together with the acquired controllers, also transfer to a physical quadrupedal manipulator without real-robot training or fine-tuning, achieving 18/20 successes in rack retrieval and 16/20 in floor-to-rack placement.
Videos and demonstrations are available at \url{https://agentic-quad-elevate.github.io/public/}.
\end{abstract}

\suppressfloats[t]
\begin{figure}[t]
    \centering
    \includegraphics[width=\columnwidth]{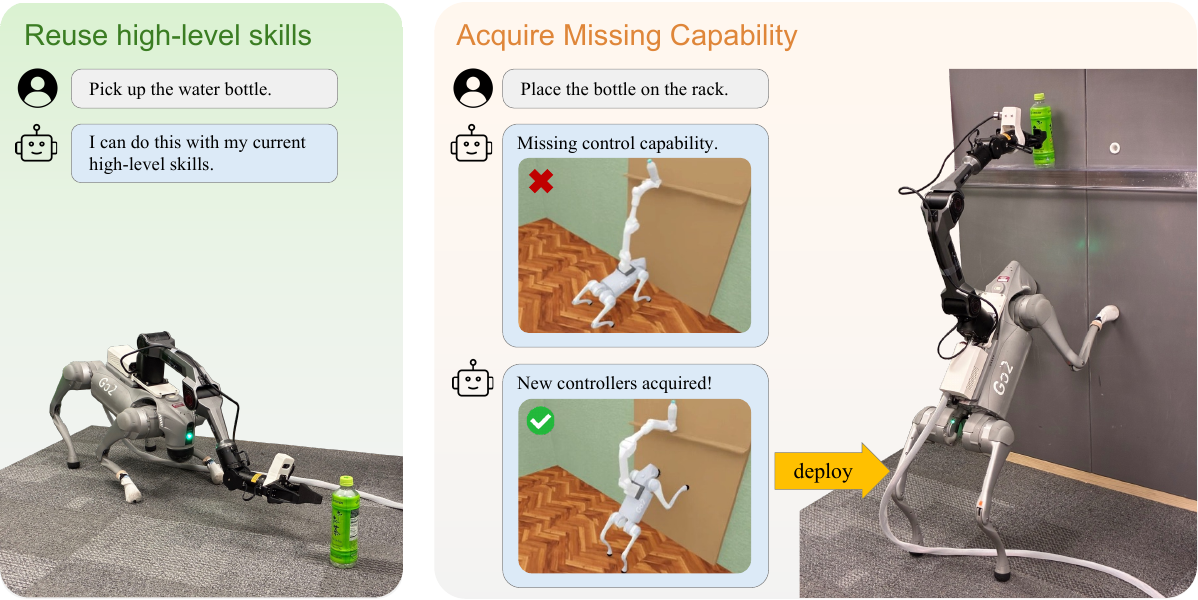}
    \vspace{-2em}
    \caption{\textbf{Reusing existing controllers and learning missing capabilities.}
    The robot uses its initial loco-manipulation controller to pick up a bottle. When a
    wall-mounted rack lies beyond its reach, it identifies the capability gap
    and trains wall-supported controllers in simulation to complete the task.}
    \label{fig:overview}
    \vspace{-1em}
\end{figure}

%% file: sessions/intro.tex
\section{Introduction}

Quadrupedal manipulators combine mobility with rich whole-body interaction capabilities~\cite{fu2023deepwholebody}. Beyond walking and manipulating objects with an arm, their morphology allows them to rear up, brace against the environment, use their legs for manipulation, and exploit contact with surrounding structures to extend their workspace~\cite{cheng2023legmanip,ouyang2025longhorizon}. These capabilities make quadrupedal manipulators promising platforms for operating in complex real-world environments~\cite{qiu2025wildlma}, but also create a fundamental challenge: the set of useful behaviors can be difficult to anticipate in advance. A task that appears impossible under the robot's current controller may become feasible once the robot discovers a different mode of physical interaction. For example, a target mounted high on a wall may be unreachable using a standard loco-manipulation controller, yet accessible if the robot first places its front feet on the wall and raises its body.

Recent advances in foundation-model-based robotics enable flexible task planning, program synthesis, and skill composition~\cite{liang2023codeaspolicies,xu2023robotool,ichter2023saycan,fu2026capx,lu2026aspire}. However, composing existing skills remains bounded by the physical capabilities of the underlying controllers.
This limitation is especially important for legged mobile manipulators. In many tabletop manipulation settings, a generic end-effector controller provides a sufficiently expressive interface for composing a broad range of tasks. A quadrupedal manipulator, in contrast, may need qualitatively different modes of whole-body control for locomotion, rearing, bipedal motion, wall bracing, leg manipulation, or contact-rich arm manipulation. Prior work has therefore developed specialized learned controllers for different behaviors~\cite{cheng2023legmanip,ouyang2025longhorizon}. For example, some systems combine locomotion with wall manipulation~\cite{cheng2023legmanip}, while others use separate locomotion, bipedal, and manipulation policies to solve long-horizon tasks~\cite{ouyang2025longhorizon}. Although such controller libraries can support diverse behaviors, the choice of which controllers to pretrain remains task-dependent. Future tasks may require capabilities that were not anticipated when the library was constructed.

A plausible alternative is to automate controller acquisition. Existing methods automate reward and training design for human-specified behaviors~\cite{ma2024eureka,ma2024dreureka}, or propose tasks and acquire skills from environmental context~\cite{wang2024robogen,zhao2026agenticskill}. However, a key challenge remains: how can an agent determine which new whole-body capability is needed when its existing controllers cannot solve a task? We study how persistent failures during task-program refinement can guide an agent to identify a missing reusable capability, formulate its reinforcement learning (RL) problem using supplied training templates, and integrate the learned controller to retry the original task, without a human specifying the new behavior to learn.

We present \method{}, a framework that uses task failures to guide the acquisition of new whole-body capabilities for quadrupedal manipulation. Following the task-program refinement pipeline used in ASPIRE~\cite{lu2026aspire}, the agent iteratively refines task programs using its available skills. When persistent failures suggest a missing physical capability, it learns a new controller through reinforcement learning and composes it with existing controllers to resume task solving. The acquired controllers and composed skills are retained for subsequent tasks, allowing the robot's capability library to grow with task demands.

We demonstrate this framework through autonomous controller acquisition and composition for elevated reaching, posture transitions, and object manipulation on a quadrupedal manipulator. The agent identifies missing capabilities, constructs their reinforcement-learning problems, trains the corresponding policies, and reuses previously acquired behaviors when solving subsequent tasks. We further validate the acquired capabilities by deploying the controllers without additional policy training for rack retrieval and floor-to-rack placement on a physical robot.

Our contributions are:
\begin{itemize}
\item \textbf{Identifying missing capabilities from task failures.} We introduce a framework that uses persistent task-program failures to identify missing whole-body capabilities, without a human specifying the new behavior to learn.
\item \textbf{Automatically training the required controllers.} The agent uses reusable training templates to design and iteratively refine rewards, curricula, and termination conditions for acquiring the identified capabilities.
\item \textbf{Reusing learned capabilities across tasks.} The framework retains learned controllers and composed skills for subsequent tasks. We demonstrate acquisition and reuse across seven simulation tasks and validate the acquired capabilities on a physical quadrupedal manipulator without additional policy training.
\end{itemize}

%% file: sessions/related_work.tex
\section{RELATED WORK}
\label{sec:related_work}

\subsection{Foundation Models for Planning and Skill Acquisition}

Foundation models enable language-based planning, program synthesis, and skill composition over robot interfaces~\cite{ichter2023saycan,xu2023robotool,fu2026capx}. Code-as-Policies~\cite{liang2023codeaspolicies} generates programs that invoke APIs for perception and control, while ASPIRE~\cite{lu2026aspire} iteratively refines task programs from execution feedback and retains reusable Python skills. These methods expand task-level behavior by recomposing available interfaces, but task execution remains bounded by the physical capabilities exposed by the underlying controllers.

A complementary line of work uses foundation models to automate skill acquisition. Text2Reward~\cite{xie2024text2reward} and Eureka~\cite{ma2024eureka} generate reinforcement-learning rewards, while DrEureka~\cite{ma2024dreureka} additionally designs domain randomization for sim-to-real transfer. RoboGen~\cite{wang2024robogen} generates tasks, environments, and training supervision, and RL-GPT~\cite{liu2024rlgpt} integrates coded actions into reinforcement learning in Minecraft. Agentic Skill Discovery~\cite{zhao2026agenticskill} learns and chains policies for simulated tabletop manipulation, acquiring missing skills during task decomposition.

\method{} connects these two lines of work. It begins with task-program synthesis and refinement; when persistent execution failures suggest a missing physical capability, the agent formulates and refines a whole-body controller-training problem using supplied training templates. The validated controller is then integrated through reusable Python skills so that task solving can resume with the acquired capability.

\subsection{Quadrupedal Loco-Manipulation}

Quadrupedal loco-manipulation builds on whole-body controllers that coordinate locomotion and arm motion~\cite{fu2023deepwholebody,portela2025wholebody}. Legs as Manipulator~\cite{cheng2023legmanip} uses a behavior tree to compose locomotion and leg-manipulation policies. Learned planners select pretrained controllers for goalkeeping~\cite{huang2023goalkeeper} and ball manipulation on rugged terrain~\cite{zhu2026dynamicball}. In locomotion, ANYmal Parkour~\cite{hoeller2024anymalparkour} selects obstacle-specific policies, while Yu et al.~\cite{yu2026skillswitching} continuously compose frozen gait and recovery policies through a learned gating network. At the task level, Ouyang et al.~\cite{ouyang2025longhorizon} generate programs over locomotion and manipulation skills and fine-tune policies using predecessor terminal states to improve transitions, while WildLMa~\cite{qiu2025wildlma} uses a large language model (LLM) planner to compose visuomotor skills and analytical navigation routines above a shared whole-body controller.

These systems demonstrate the value of composing specialized behaviors from prepared repertoires. \method{} studies how task-execution failures can motivate expansion of the repertoire itself: the agent hypothesizes a missing whole-body behavior, trains and validates a corresponding controller, and retains it for composition and reuse in later tasks.

%% file: sessions/methodology.tex
\section{Methodology}
\label{sec:methodology}

\begin{figure*}[t]
    \centering
    \includegraphics[width=\textwidth]{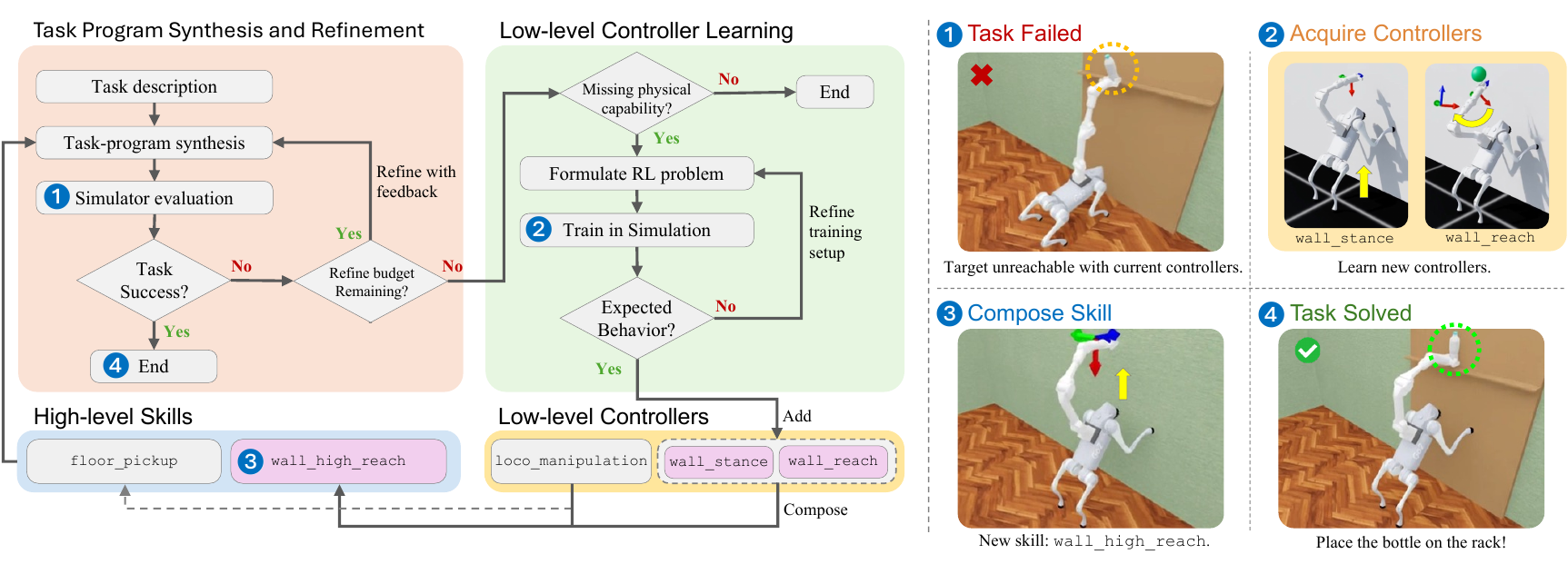}
    \vspace{-2.5em}
    \caption{\textbf{Overview of \method{}.} The agent synthesizes and refines task programs using its capability library. Persistent failures after refinement (1) motivate controller learning (2): the agent assesses feasibility, formulates an RL problem, and revises training using behavior feedback. Validated controllers are added to the library and used to develop skills (3), enabling renewed task execution and eventual completion (4). The right panels schematically illustrate bottle placement using Run~5's controller and skill names: \texttt{wall\_stance} elevates the body using wall support, while \texttt{wall\_reach} enables manipulation from that stance. The \texttt{wall\_high\_reach} skill uses these controllers to reach elevated targets. Feasibility and behavior checks reflect the agent's assessments; the reported acquisition sequence is described in Sec.~\ref{sec:exp_wall}.}
    \label{fig:method}
    \vspace{-1.5em}
\end{figure*}

\method{} enables a quadrupedal manipulator to learn new controllers in response to persistent task failures. Given a language instruction and a simulation environment, the agent first attempts the task using its existing controllers and skills. It refines the task program based on feedback from development trials. When persistent failures suggest a missing physical capability, the agent formulates and refines a reinforcement-learning problem to train a new controller. It then integrates the new controller into task programs alongside existing ones and retries the original task. Figure~\ref{fig:method} summarizes the three stages: task refinement and capability diagnosis, controller formulation and training, and controller integration and skill reuse.

A \emph{task program} is executable Python code for a particular task; it can invoke skills and include task-specific logic. Task programs use a capability library containing two types of reusable components: controllers and skills. A \emph{controller} is a learned policy that maps numerical observations and supported commands to joint actions. A \emph{skill} is a reusable Python function that packages a task-level robot behavior using available controllers and environment feedback, such as reaching a target or picking up an object. \emph{Helper functions} provide supporting operations for skills and task programs, including perception estimates, measurements, geometric calculations, base navigation, and end-effector pose control. Both skills and helpers can issue controller commands; skills combine supporting operations into reusable task-level behaviors.

\begin{figure}[t]
    \centering
    \includegraphics[width=0.95\columnwidth]{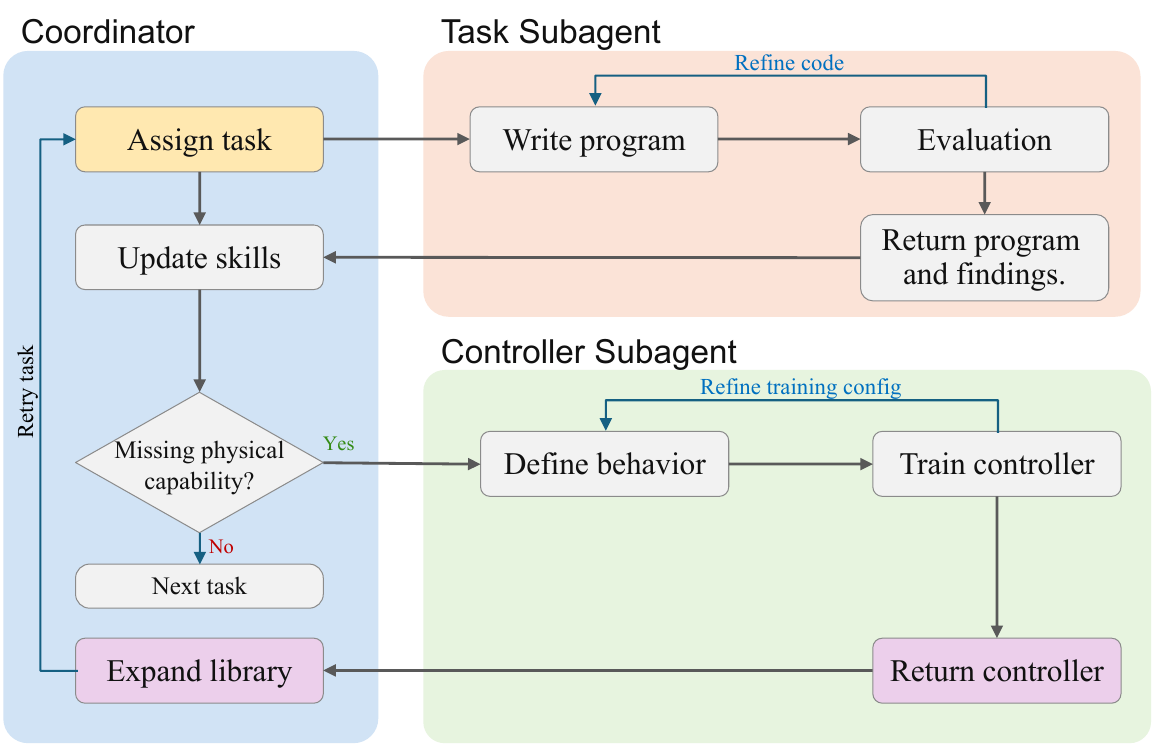}
    \caption{\textbf{Agent workflow.} Task subagents refine programs using development feedback. When failures suggest a missing physical capability, the coordinator dispatches a controller subagent to formulate and refine training. Validated controllers expand the library, enabling a fresh task attempt.}
    \label{fig:app_workflow}
    \vspace{-1em}
\end{figure}

The library initially contains a pretrained loco-manipulation controller (\texttt{loco\_manipulation}) for base motion and 6D end-effector pose control, together with the supplied \texttt{floor\_pickup} skill, which uses this controller for visually guided object pickup from the floor. The agent expands this library by training new controllers and developing reusable skills. Figure~\ref{fig:app_workflow} summarizes how the coordinator and subagents collaborate in this process.

\subsection{Task Refinement and Capability Diagnosis}
\label{sec:method_program}

Given a task, \method{} first generates and refines a Python program that uses its available controllers and skills. A coordinator assigns this process to a task subagent in a fresh context. Following ASPIRE's exploration-and-repair loop~\cite{lu2026aspire}, the subagent inspects an initial observation, generates a Python program that invokes available skills, and tests it across development variants. Between executions, it uses numerical outcomes, execution traces, and visual feedback to refine the task program. Each failed variant receives a bounded number of repair replays. Controller weights remain fixed throughout program refinement.

When program refinement repeatedly fails, the task subagent assesses whether the task requires a physical capability missing from its current controller library. A capability gap is suggested when different valid strategies encounter the same physical limitation, such as consistently falling short of an elevated target despite successful approach motions. The task subagent records the attempted strategies and remaining errors to support this diagnosis.

When failures across most development variants suggest a missing physical capability, the task subagent reports the suspected capability gap and supporting evidence to the coordinator. The coordinator reviews the report and, if acquisition is warranted, dispatches a controller subagent. This subagent identifies a reusable behavior to learn and formulates and refines a reinforcement-learning problem to train the corresponding controller. Acquisition decisions use only development feedback; Appendix~\ref{app:pipeline} details the full procedure.

\subsection{Controller Formulation and Training}
\label{sec:method_learning}

Given the diagnosed capability gap, the controller subagent identifies the behavior to learn and designs the reinforcement-learning setup needed to acquire it. It inspects robot geometry and training code to assess feasibility, then specifies the training scene, supported commands, and initial states. It designs behavior-specific rewards, curricula, termination conditions, and measurable success criteria, and defines when control transfers from the preceding controller. The framework supplies the robot model, simulator, proximal policy optimization (PPO) trainer~\cite{schulman2017ppo}, and reusable training templates. Table~\ref{tab:autonomy_scope} summarizes this division of responsibility.

\begin{table}[t]
    \centering
    \caption{Division of responsibility in \method{}.}
    \label{tab:autonomy_scope}
    \footnotesize
    \setlength{\tabcolsep}{3pt}
    \renewcommand{\arraystretch}{1.0}
    \begin{tabular}{>{\raggedright\arraybackslash}p{0.22\columnwidth}>{\raggedright\arraybackslash}p{0.34\columnwidth}>{\raggedright\arraybackslash}p{\dimexpr0.44\columnwidth-6\tabcolsep\relax}}
        \toprule
        \textbf{Stage} & \textbf{Provided by framework} & \textbf{Agent decisions and outputs} \\
        \midrule
        Capability identification & Initial capability library; public task interface & Failure diagnosis; reusable behavior to acquire \\
        \addlinespace[2pt]
        \rowcolor{black!7}
        Training formulation & Two behavior templates; command modes; posture and support guidance & Scene and reset configurations; rewards, command distributions, curricula, terminations, and success criteria \\
        \addlinespace[2pt]
        Training and validation & Robot model, simulator, PPO settings, shared regularization and randomization & Revisions based on training feedback; training execution; checkpoint selection and validation \\
        \addlinespace[2pt]
        \rowcolor{black!7}
        Integration and reuse & Controller registry; guarded switching; bounded target correction & Switch conditions; reusable skills; controller composition and library updates \\
        \bottomrule
    \end{tabular}
    \vspace{-2em}
\end{table}

Each controller is a policy $u_t = \pi_{\theta}(o_t,c_t)$ with learned parameters $\theta$. The \emph{controller observation} $o_t$ contains robot and environment feedback and action history, while the \emph{controller command} $c_t$ specifies supported targets or modes. Supported manipulation encodes the pose command as keypoint errors in the measured end-effector frame, together with an arm-motion flag (Appendix~\ref{app:training_settings}). These state-relative targets remain commands. Together, $(o_t,c_t)$ form the \emph{controller input}; for a controller without commands, the input consists of $o_t$ alone. The policy outputs an 18-dimensional action vector $u_t$. The runtime converts it to joint-position targets $q_t^{\mathrm{tgt}} = q^{\mathrm{default}} + 0.25\,u_t$, where $q^{\mathrm{default}}$ contains the controller-specific default positions for the 12 leg and 6 arm joints. Controllers share this action space, while their supported commands depend on the behavior.

We provide two templates to guide behavior-specific controller training.
The \emph{body reconfiguration} template trains a controller to move the robot into a new posture and hold it. The template also includes reward terms that encourage the arm to remain near its gravity-compensated home pose for balance. The controller subagent chooses the support structure and target posture, then uses the supplied guidance to design rewards for posture progress, support contacts, and stability. The template provides optional temporary pose assistance to help the robot reach the target posture selected by the controller subagent. Assistance may be used in diagnostic evaluations during training, but is disabled for final controller validation and task evaluation.

The \emph{supported manipulation} template trains a controller to track commanded end-effector poses while maintaining a supported stance. During arm motion, the robot is encouraged to keep its leg-joint configuration and base pose close to their initial values. Training targets are sampled from a dataset of reachable end-effector poses, which we construct by sampling arm configurations within joint limits and recording the resulting end-effector poses in simulation.

After selecting a training template, the controller subagent initializes training episodes from states recorded during rollouts of the preceding controller. This allows the new controller to learn from states that its predecessor can produce. The subagent then runs configuration and simulation checks, conducts a short PPO pilot, and trains and evaluates candidate controllers. Numerical measurements and recorded rollouts guide revisions to rewards, training conditions, and termination thresholds. Inherited termination rules are adjusted when they conflict with the intended posture, while criteria for falls and undesired contacts are retained. The subagent trains or resumes the revised candidate and checks whether it achieves the intended support contacts, body configuration, or reaching behavior. This validation precedes integration into task programs. Shared regularization, PPO settings, domain randomization, and training budgets are supplied by the framework.

\subsection{Controller Integration and Skill Reuse}
\label{sec:method_reuse}

After completing training and validation, the controller subagent reports the results to the coordinator, which adds the new controller to the capability library. Task programs switch controllers only when the target controller's switch conditions are satisfied; implementation details are provided in Appendix~\ref{app:controller_switching}.

One task may require successive controller acquisitions. A prerequisite controller is validated on its intended behavior without requiring completion of the original task. For example, a body reconfiguration controller can raise the robot into a wall-supported posture, followed by a supported manipulation controller for reaching from that posture. Once the required controllers are available, the coordinator dispatches a fresh task subagent to retry the original task using the expanded library. The task subagent refines the program and develops skills for reusable robot behaviors using the available controllers. Controller weights remain fixed during this process.

Following ASPIRE~\cite{lu2026aspire}, task subagents identify reusable task behaviors during program refinement and report them to the coordinator. The coordinator retains these behaviors as parameterized skills, with documentation of their applicability, usage, and limitations. Both learned controllers and reusable skills remain in the library for subsequent tasks, while the environment resets between tasks.

%% file: sessions/experiments.tex
\section{Experiments}
\label{sec:experiments}

Our experiments evaluate whether \method{} can autonomously expand a quadrupedal manipulator's capabilities in response to task failures within the supplied controller-training framework. Specifically, we ask whether the agent can
(i) identify missing control capabilities from task failures, 
(ii) acquire usable controllers to address those failures, and
(iii) reuse them on later tasks.
We additionally validate the acquired capabilities by deploying the resulting controllers on a physical robot without policy fine-tuning.

\setcounter{dbltopnumber}{1}
\begin{figure*}[t]
    \centering
    \includegraphics[width=\textwidth]{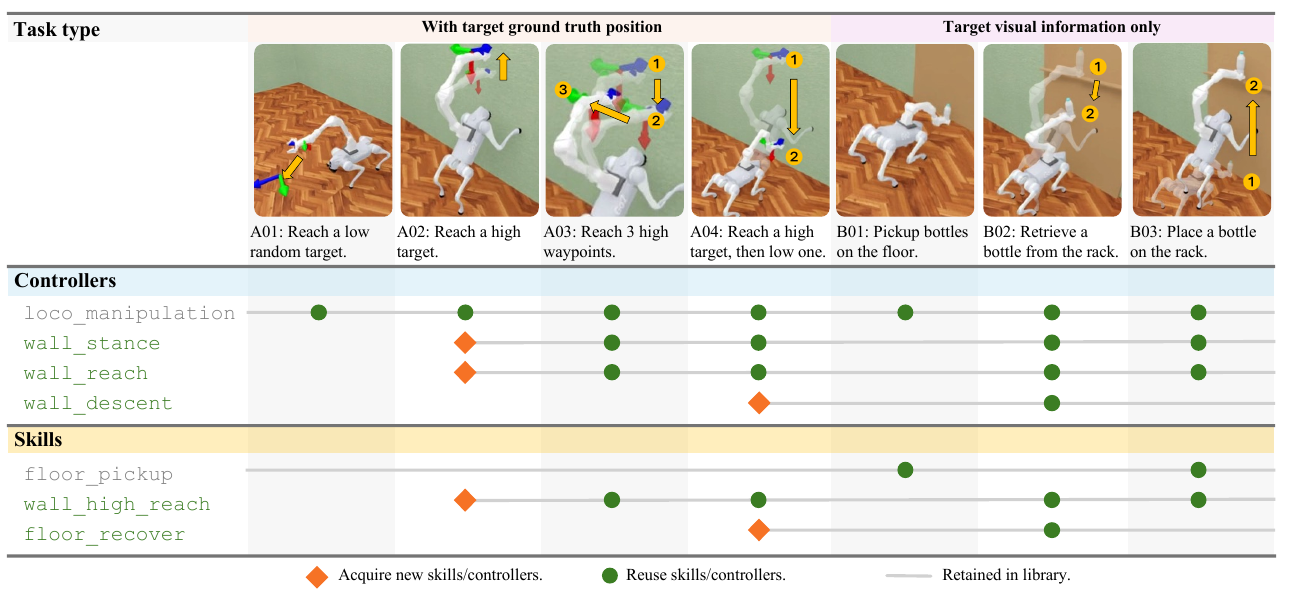}

    \vspace{-1em}
    \caption{\textbf{Controller acquisition and reuse across tasks in Run~5.} Tasks proceed left to right with the library retained. A01 and B01 use the initial controller; the other five tasks require capabilities beyond its workspace. Diamonds mark controller acquisition or skill creation, circles mark use, and gray lines indicate continued availability. Table~\ref{tab:task_success} reports task success rates across all five runs. B01 includes bottles and soda cans.}
    \label{fig:task_sequence}

    \vspace{-1.5em}
\end{figure*}

\subsection{Experimental Setup}
\label{sec:exp_setup}

The task suite comprises four reaching tasks (A01--A04) followed by three object-manipulation tasks (B01--B03). 
It progresses from floor-level reaching to elevated targets and posture transitions, then tests whether the accumulated capabilities support object manipulation. 
A01 and B01 are within the initial controller library capability; 
A02--A04 and B02--B03 require elevated reaching beyond its workspace. We conduct five independent runs using Claude Opus~5.5 through the Claude Code harness. Figure~\ref{fig:task_sequence} illustrates the tasks and the reuse of controllers and skills in Run~5; Table~\ref{tab:task_suite} defines each task and its success criterion.

We use a Unitree Go2 quadruped with 12 leg joints and a 6-DoF AgileX PiPER arm. The robot carries a built-in front camera and LiDAR on the base and an Intel D405 RGB-D camera at the end effector (Fig.~\ref{fig:tasks_real}). Task programs operate at 10~Hz, controllers at 50~Hz, and simulation physics at 200~Hz.
Task programs access proprioceptive observations comprising joint positions and velocities and IMU-based estimates of base orientation. Object segmentation masks are provided by Isaac Lab~\cite{mittal2025isaaclab} in simulation and SAM~3~\cite{carion2025sam3} on the physical robot. We use the front LiDAR only to estimate the robot’s distance and heading relative to the surface ahead. A01--A04 additionally provide end-effector-to-target pose errors, while B01--B03 rely on vision for target localization and receive no target-position observations. The physical perception pipeline is described in Sec.~\ref{sec:exp_real}.

\begin{table}[t]
    \centering
    \caption{Task suite and success criteria.}
    \label{tab:task_suite}
    \scriptsize
    \setlength{\tabcolsep}{3pt}
    \renewcommand{\arraystretch}{1.0}
    \begin{tabular}{@{}>{\raggedright\arraybackslash}p{0.06\columnwidth}>{\raggedright\arraybackslash}p{\dimexpr0.94\columnwidth-2\tabcolsep\relax}@{}}
        \toprule
        Task & Objective and success criterion \\
        \midrule
        A01 & Reach a random floor-level target. \\
        \addlinespace[2pt]
        \rowcolor{black!7}
        A02 & Reach a target 1.25~m above the floor, 0.30~m from the wall. \\
        \addlinespace[2pt]
        A03 & Reach three elevated waypoints in order. \\
        \addlinespace[2pt]
        \rowcolor{black!7}
        A04 & Reach targets at 1.25~m then 0.40~m in the same wall column. \\
        \midrule
        B01 & Lift a floor bottle or soda can at least 0.30~m above the floor. \\
        \addlinespace[2pt]
        \rowcolor{black!7}
        B02 & Retrieve a bottle or can from a 1.05~m-high rack; stop upright with four feet on the floor, arm homed, and object held. \\
        \addlinespace[2pt]
        B03 & Pick up a floor bottle or can and place it upright on the rack. \\
        \bottomrule
    \end{tabular}
    \par\smallskip
    \begin{minipage}{\columnwidth}
        \footnotesize
        All success conditions must hold continuously for 1~s. For A01--A04, hold each target pose within 0.05~m position and 0.20~rad orientation error.
    \end{minipage}
    
    \vspace{-1em}
\end{table}

Tasks run in the order A01 $\rightarrow$ A02 $\rightarrow$ A03 $\rightarrow$ A04 $\rightarrow$ B01 $\rightarrow$ B02 $\rightarrow$ B03. The initial library contains a pretrained loco-manipulation controller for base motion and 6D end-effector pose control, together with the supplied \texttt{floor\_pickup} Python skill and perception and manipulation notes. The environment resets between tasks, while previously acquired controllers and skills remain available.

We compare \method{} with Code-as-Policies~\cite{liang2023codeaspolicies} and ASPIRE~\cite{lu2026aspire} to evaluate how acquiring new controllers expands the range of tasks the robot can solve. We implement both baselines as controlled variants in the same framework: ASPIRE uses \method{}'s task-program synthesis, feedback-based repair, and reusable-skill accumulation with controller acquisition disabled. Code-as-Policies additionally disables program repair, executing one initial program per task unchanged across variants. Both baselines retain the initial low-level controller throughout the sequence; \method{} can train and add controllers when development failures indicate a missing capability.

All three methods use the same language model and reasoning setting, initial capabilities, task order, task observations, evaluation variants, success rules, and episode limits. \method{} and ASPIRE share the development-selection rule and allow at most three repair replays per failed development variant; Code-as-Policies allows none. Controller training and the subsequent program development enabled by library expansion are additional computation used by \method{}; total computation is therefore not matched across methods.

Each simulation task contains 15 development variants and 50 separate evaluation variants. Reaching variants vary target poses or initial robot poses; object-manipulation variants vary object placement. Each variant is fixed across resets and retries. The coordinator runs held-out evaluation; subagents never receive its results. Evaluation outcomes are not used for task-program refinement or controller acquisition.

Task programs, reusable library code, and controller weights remain fixed throughout held-out evaluation. The seven-task aggregate includes the final 50 trials per task in each of five independent runs (1,750 trials per method). We report mean success rates and sample standard deviations across the five run-level rates; detailed protocols and per-run outcomes appear in Appendices~\ref{app:evaluation_protocol} and~\ref{app:independent_runs} and Table~\ref{tab:app_baseline_runs}.

\subsection{Task Performance and Coverage}
\label{sec:exp_comparison}

\begin{table}[t]
    \centering
    \caption{\textbf{Task success rates (\%).} Mean $\pm$ sample standard deviation across five independent runs, 50 variants per task.}
    \label{tab:task_success}
    \footnotesize
    \setlength{\tabcolsep}{2pt}
    \renewcommand{\arraystretch}{1.0}
    \begin{tabular*}{\columnwidth}{@{\extracolsep{\fill}}l*{3}{r}@{}}
        \toprule
        Task & Code-as-Policies & ASPIRE & \method{} \\
        \midrule
        A01 & $73.2 \pm 11.6$ & $\mathbf{99.6 \pm 0.9}$ & $99.2 \pm 1.1$ \\
        A02 & $0.0 \pm 0.0$ & $0.0 \pm 0.0$ & $\mathbf{100.0 \pm 0.0}$ \\
        A03 & $0.0 \pm 0.0$ & $0.0 \pm 0.0$ & $\mathbf{98.4 \pm 2.6}$ \\
        A04 & $0.0 \pm 0.0$ & $0.0 \pm 0.0$ & $\mathbf{100.0 \pm 0.0}$ \\
        \addlinespace[2pt]
        B01 & $88.0 \pm 18.3$ & $\mathbf{99.2 \pm 1.8}$ & $97.6 \pm 3.6$ \\
        B02 & $0.0 \pm 0.0$ & $0.0 \pm 0.0$ & $\mathbf{99.6 \pm 0.9}$ \\
        B03 & $0.0 \pm 0.0$ & $0.0 \pm 0.0$ & $\mathbf{98.4 \pm 0.9}$ \\
        \midrule
        Overall & $23.0 \pm 3.4$ & $28.4 \pm 0.3$ & $\mathbf{99.0 \pm 0.7}$ \\
        \bottomrule
    \end{tabular*}
    \vspace{-2em}
\end{table}

On A01 and B01, which are supported by the initial controller, \method{} shows no advantage over ASPIRE (Table~\ref{tab:task_success}). The coverage gain occurs on A02--A04 and B02--B03, where both baselines with a fixed controller library fail on every held-out trial and \method{} achieves 98.4\%--100\% mean success. These results reflect the task suite's deliberate inclusion of missing physical capabilities and do not establish superior high-level program synthesis.

Across the seven tasks and five runs, \method{} achieves 99.0\% overall success, compared with 28.4\% for ASPIRE and 23.0\% for Code-as-Policies. This comparison includes additional controller training and subsequent program development only for \method{}. Acquiring the required working postures resolves the observed workspace limitation while leaving execution failures within the expanded workspace.

\subsection{Autonomous Controller Acquisition}
\label{sec:exp_wall}

On A02 in Run~5, repeated development sweeps with the initial controller yield 0/15 successes. The best revised program reduces vertical error from 0.22--0.24~m to 0.20--0.21~m, still about four times the 0.05~m tolerance. The agent diagnoses insufficient arm-base elevation and proposes \texttt{wall\_stance} for wall-supported elevation and \texttt{wall\_reach} for end-effector tracking from that stance.

Using the supplied templates and PPO infrastructure, the agent autonomously specifies and revises environment configurations, rewards, termination conditions, and curricula, and selects checkpoints based on training and evaluation feedback.

\begin{figure}[t]
    \centering
    \includegraphics[width=0.85\columnwidth]{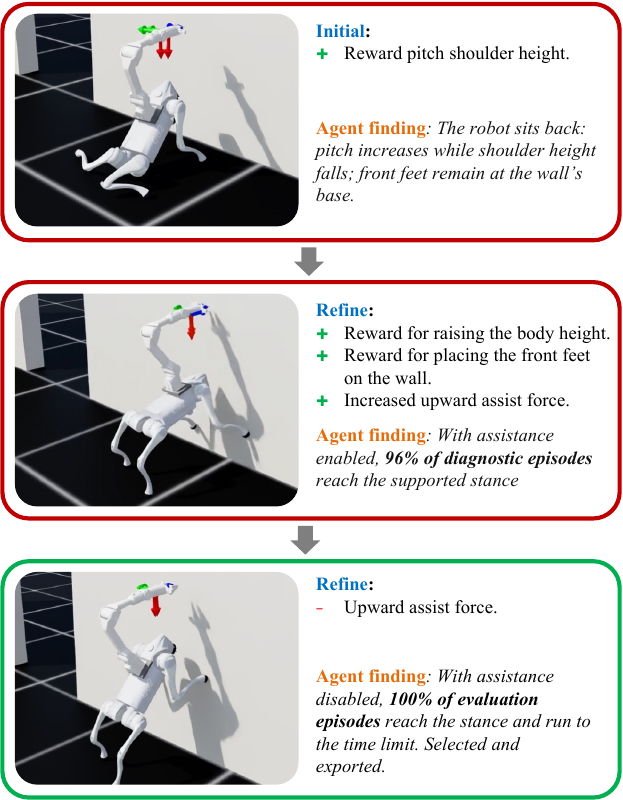}
    \vspace{-1em}
    \caption{\textbf{Autonomous improvement through reward revision on A02 in Run~5.} Top: the initial controller sits backward without the intended elevation and wall support. Middle: after reward revisions and stronger upward assistance, 96\% of diagnostic episodes reach the stance with assistance enabled. Bottom: after further training, 100\% of evaluation episodes reach the stance with assistance disabled.}
    \label{fig:a02_acquisition}
    \vspace{-1em}
\end{figure}

Training \texttt{wall\_stance} illustrates feedback-driven revision of the training formulation (Fig.~\ref{fig:a02_acquisition}). The initial policy sits backward, increasing pitch without sufficient body elevation or wall support. After inspecting reward components and recorded rollouts, the agent revises the formulation to encourage elevation and appropriate support contacts. Following these changes and further training, the selected controller establishes the intended stance in evaluations with simulated body assistance disabled.

On A04 in Run~5, the agent learns a \texttt{wall\_descent} controller for returning from wall support to the floor. The \texttt{floor\_recover} skill combines it with \texttt{loco\_manipulation} to restore a stable four-foot stance and resume standing loco-manipulation, enabling the robot to reach the low target.

In all five reported runs, \method{} acquires wall-supported elevation and supported manipulation on A02, followed by wall descent on A04. Each run starts without controllers, skills, or findings acquired by another run. In every run, A02 and A04 each improve from 0/50 held-out successes before acquisition to 50/50 afterward.
Total controller training uses 4.2--6.3~h of computation per run on one NVIDIA RTX 5090, summing rollout-collection and learning times across pilots, restarts, and resumed segments. Appendix~\ref{app:compute} reports the per-run effort and full workflow timing.
All five runs use the same initial repository configuration, task interfaces, task success rules, held-out variants, and training budgets. Appendix~\ref{app:independent_runs} details these conditions and the individual outcomes.

\subsection{Different Environment, Different Strategies}
\label{sec:exp_support}

\begin{figure}[t]
    \centering
    \includegraphics[width=0.85\columnwidth]{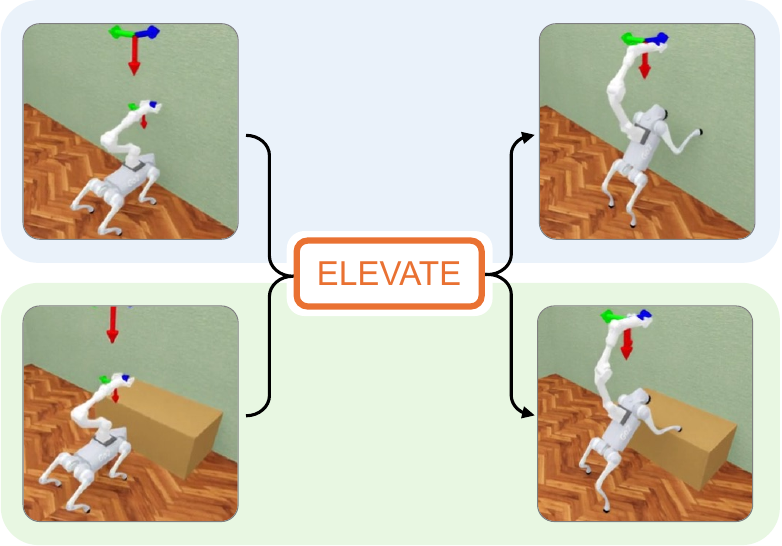}
    \caption{\textbf{Environment-specific support strategies.} Initial scenes (left) and acquired behaviors (right) for elevated reaching. The agent learns front-foot support against the wall (top) or on an added box (bottom), with the same initial controller library and task instructions.}
    \label{fig:sim_diverse}
    \vspace{-1em}
\end{figure}

In a qualitative variant of A02 with an added box, \method{} acquires two controllers: \texttt{box\_stance} for front-foot support on the box and \texttt{box\_reach} for end-effector tracking from that stance (Fig.~\ref{fig:sim_diverse}). The initial controller library and task instructions remain unchanged, and the agent is not instructed to use the box. Together with the wall-supported behavior, this example illustrates how the agent can select an available support structure when acquiring a capability for elevated reaching.

\setcounter{dbltopnumber}{2}
\begin{figure*}[t]
    \centering
    \includegraphics[width=\textwidth]{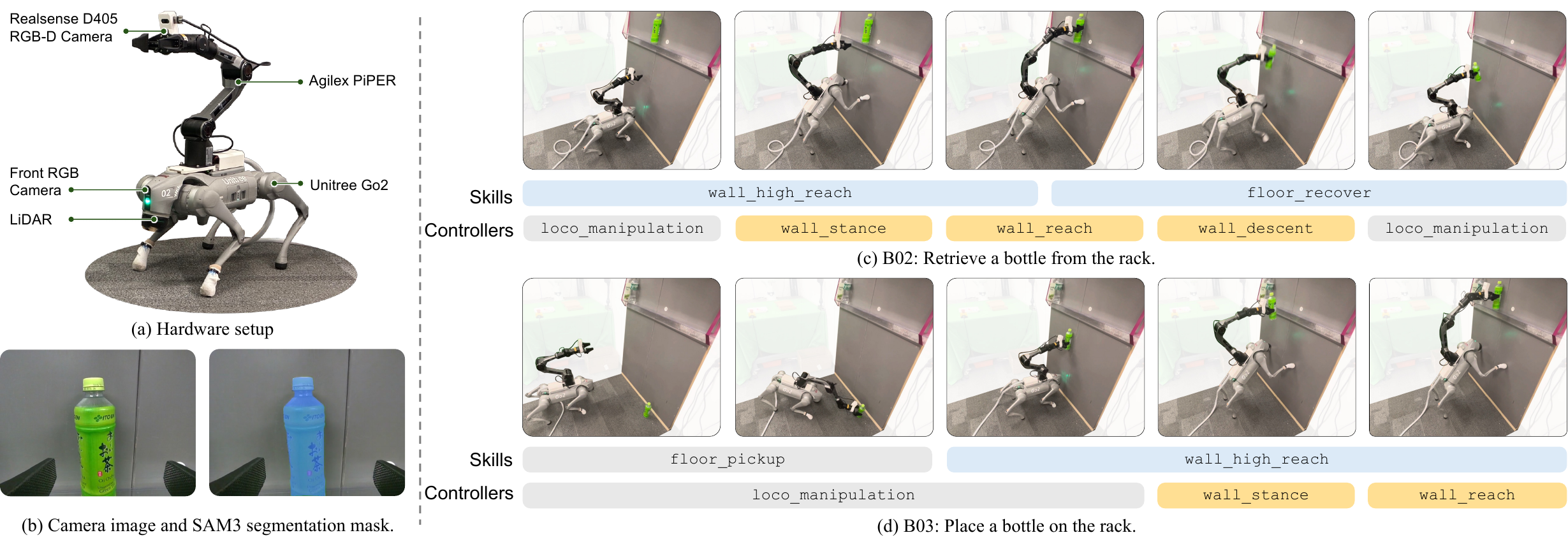}

    \vspace{-1em}
    \caption{\textbf{Hardware, perception, and zero-shot physical deployment.} (a) Unitree Go2 with a 6-DoF AgileX PiPER arm, Intel D405 RGB-D camera, front RGB camera, and LiDAR. (b) Camera image and corresponding SAM~3 segmentation mask. (c) B02: rack retrieval and return to a four-foot stance while retaining the bottle. (d) B03: floor pickup and upright rack placement. Bands show active skills and controllers. With unchanged controller weights from simulation, B02 and B03 achieve 18/20 (90\%) and 16/20 (80\%) successes, respectively.}
    \label{fig:tasks_real}
    \vspace{-1.5em}
\end{figure*}

\subsection{Capability Reuse Across Tasks}
\label{sec:exp_reuse}

In all five runs, acquisition on A02 and A04 expands the library from one to four controllers, with no additional controller training on A03 or B01--B03. The following examples illustrate reuse in Run~5 (Fig.~\ref{fig:task_sequence}); Appendix~\ref{app:run_artifacts} lists each run's controllers, skills, and helpers.

On A03, the two controllers acquired at A02 are reused unchanged without additional controller training. Its program adapts the \texttt{wall\_high\_reach} skill to visit three elevated waypoints, demonstrating reuse for a longer reaching sequence.

Reuse also extends from reaching to object manipulation. B02 uses the \texttt{wall\_high\_reach} skill for elevated retrieval, then reuses \texttt{floor\_recover} to return to the floor while retaining the bottle. B03 combines floor pickup with the elevated-reaching skill to transport and place the bottle on the rack. No additional controllers are trained in B01--B03. These executions extend the use of the acquired controllers from reaching to retrieval and placement through task-specific perception and manipulation code.

\subsection{Zero-Shot Deployment on the Physical Robot}
\label{sec:exp_real}

We deploy the acquired controllers on the physical robot for B02 and B03, which combine floor-level and elevated manipulation (Fig.~\ref{fig:tasks_real}). B02 requires retrieval and a return to a settled four-foot stance while retaining the bottle; B03 requires floor pickup and upright placement on the rack. Across the two tasks, the programs compose controllers for loco-manipulation, wall-supported elevation, supported manipulation, and wall descent. All controller weights transfer unchanged from simulation, with no training or fine-tuning on real-robot data.

Physical execution uses image-based perception and the real gripper. SAM~3~\cite{carion2025sam3} supplies object masks. The Go2's front camera supports search and approach, while masks from the end-effector D405 camera are combined with aligned depth and calibration to localize manipulation targets in the robot frame. Task programs use these estimates to coordinate approach, grasping, controller transitions, and placement.

Over 20 trials per task, B02 achieves 18/20 successes (90\%) and B03 achieves 16/20 (80\%). In the two B02 failures, limited gripper--bottle contact allows the bottle to slip during the large accelerations of floor recovery. B03 failures mainly arise from unintended gripper contact during floor pickup. These trials demonstrate physical execution of the composed capabilities without policy fine-tuning, while identifying grasp robustness during pickup and body transitions as a remaining limitation.

%% file: sessions/conclusion.tex
\section{Conclusion}
\label{sec:conclusion}

We presented \method{}, a framework that uses persistent task-execution failures to drive the acquisition of missing whole-body control capabilities for quadrupedal manipulation. When execution feedback suggests a missing physical capability, \method{} formulates and refines a reinforcement-learning problem using supplied training templates, validates the resulting controller, and integrates it into task programs alongside existing controllers. Across the evaluated seven-task sequence, the controller library grows from one pretrained loco-manipulation controller to four controllers. The acquired capabilities enable 98.4\%--100\% mean success across five independent runs on five tasks beyond the initial controller's workspace and are reused across subsequent reaching and object-manipulation tasks. Zero-shot deployment of the resulting controllers achieves 18/20 successes on rack retrieval and 16/20 on floor-to-rack placement on a physical quadrupedal manipulator.

%% file: sessions/references.tex
\bibliographystyle{IEEEtran}
\bibliography{references}

%% file: sessions/appendix.tex
\onecolumn
\providecommand{\apptodo}[1]{\textcolor{red}{[TODO: #1]}}
\providecommand{\ctrlloco}{\texttt{loco\_manipulation}}
\providecommand{\ctrlrear}{\texttt{wall\_stance}}
\providecommand{\ctrlreach}{\texttt{wall\_reach}}
\providecommand{\ctrldismount}{\texttt{wall\_descent}}
\newcolumntype{P}[1]{>{\raggedright\arraybackslash}p{#1}}
\useRomanappendicesfalse
\appendices

This appendix details the implementation and evaluation of \method{}. Appendix~\ref{app:agent_interfaces} presents the pipeline and scope of agent autonomy. Appendix~\ref{app:acquisition_evidence} traces controller acquisition and reuse through worked examples, and Appendix~\ref{app:controller_training} provides the training and integration specifications. Appendix~\ref{app:evaluation_deployment} describes the tasks, evaluation protocols, and baseline results. Appendix~\ref{app:run_study} compares performance, acquired capabilities, training and revision effort, runtime, and estimated model cost across the five independent Claude acquisition runs used in the main-text aggregate. Unless stated otherwise, worked examples and controller specifications refer to Claude Run~5.

\input{sessions/appendix/01_pipeline_and_autonomy.tex}

\input{sessions/appendix/02_acquisition_and_reuse.tex}
\input{sessions/appendix/03_controller_training.tex}
\input{sessions/appendix/04_experimental_protocols.tex}
\input{sessions/appendix/05_independent_runs.tex}

%% file: sessions/appendix/01_pipeline_and_autonomy.tex
\section{Pipeline and Autonomy Boundaries}
\label{app:agent_interfaces}

This section presents the workflow through which \method{} develops task programs, acquires missing controllers, and integrates them into a reusable library. We first describe the initial capabilities, task interfaces, and agent responsibilities, then explain the end-to-end procedure. Finally, we clarify which constraints the framework fixes and which decisions the agents make about behaviors, training formulations, and controller integration.

\subsection{Initial Capabilities and Task Interfaces}
\label{app:initial_instructions}

A \emph{task program} is executable Python code for a particular task; it can invoke skills and include task-specific logic. \method{} maintains two kinds of executable capability. A \emph{controller} is a learned policy that maps controller observations and supported commands to an 18-dimensional action vector, which the runtime converts to joint-position targets. A \emph{skill} is a reusable Python function that packages a task-level robot behavior using available controllers and environment feedback. Helper functions provide supporting perception, measurement, geometric, navigation, and pose-control operations for skills and task programs. Both skills and helpers can issue controller commands. The agent trains controllers during acquisition. When developing task programs, it uses the available controllers without changing their weights.

The initial library contains the frozen \ctrlloco{} controller, the \texttt{floor\_pickup} skill, and written perception and manipulation notes. The \ctrlloco{} controller tracks base-velocity commands for Go2 locomotion on the floor and 6D PiPER end-effector pose commands (position and orientation) specified in the robot's base frame. The \texttt{floor\_pickup} skill uses visual input and object segmentation masks to pick up objects from the floor by issuing commands to the \ctrlloco{} controller. The training framework supplies the robot model, simulator, PPO trainer, two controller templates, 10,000 reachable PiPER end-effector poses, and settled start states of the initial controller. Thus, the agent starts with functioning locomotion, arm control, and floor pickup, together with a structured controller-authoring framework.

\phantomsection
\label{app:runtime_interfaces}
All task programs receive proprioceptive observations, robot state, and front-surface distance and heading measurements. A01--A04 additionally provide ground-truth end-effector-to-target pose errors. B01--B03 instead provide front RGB and wrist RGB-D images, camera calibration, and segmentation masks, without ground-truth object or rack poses. These tasks therefore require the program to estimate target locations from visual observations.

\subsection{Agent Roles and Information Access}
\label{app:agent_roles}

A coordinator dispatches a fresh task subagent for each task attempt and a fresh controller subagent for each acquisition. They exchange findings and controller documentation through files rather than sharing full conversations. The coordinator publishes shared-library updates; controller subagents author controller integration code before publication. Table~\ref{tab:app_roles} summarizes the division of work and information access.

\begin{table}[!ht]
    \centering
    \caption{Agent responsibilities and information access.}
    \label{tab:app_roles}
    \footnotesize
    \setlength{\tabcolsep}{4pt}
    \begin{tabular}{@{}P{2.1cm}P{5.0cm}P{5.0cm}P{4.0cm}@{}}
        \toprule
        Role & Responsibilities & Available information & Restrictions \\
        \midrule
        Coordinator & Assigns work to subagents, publishes library updates, starts controller acquisition, approves controllers for the library, and runs held-out trials. & Development findings and reports of blocked variants; controller documentation and result summaries. & No access to raw debugging traces, keyframes, videos, or program output. Held-out results cannot guide development. \\
        \rowcolor{black!7}
        Task subagent & Develops and repairs task programs, selects the final program, and records reusable findings. & Public task and API documentation; available controllers and skills; outcomes and traces from its own development trials. & No access to private task definitions, simulator or training code, assets, datasets, or held-out variants and results. \\
        Controller subagent & Defines the capability and training design, revises training, selects a checkpoint, writes controller integration code, and runs confirmation sweeps. & Findings from blocked tasks; public interfaces; robot geometry; training and environment code; controller evaluations and development replays. & No access to private task definitions or held-out variants and results. Cannot modify the initial controller or other controller packages. \\
        \bottomrule
    \end{tabular}
\end{table}

\subsection{End-to-End Procedure}
\label{app:pipeline}

The sequence is A01 $\rightarrow$ A02 $\rightarrow$ A03 $\rightarrow$ A04 $\rightarrow$ B01 $\rightarrow$ B02 $\rightarrow$ B03. The library persists across tasks, while each execution resets the environment. The following procedure connects program repair to controller acquisition; detailed evaluation rules appear in Appendix~\ref{app:evaluation_protocol}.

\begin{enumerate}
    \item \textbf{Develop a task program.} A task subagent inspects an initial scene, states a hypothesis, and executes its initial program on the 15 development variants. It repairs failed variants with at most three further replays each, classifying exhausted variants as physical, perceptual, or algorithmic failures.
    \item \textbf{Select and freeze.} The subagent selects one program using development evidence and reports reusable patterns. The coordinator publishes the selected patterns, freezes the program and library snapshot, and runs the 50 held-out variants once. These outcomes are reported without feeding back into development.
    \item \textbf{Assess the capability gap.} After held-out evaluation is complete, the coordinator assesses the gap using only the development reports. Acquisition requires physical failures on most development variants, valid completed executions, a substantial residual error across materially different programs, and no plausible remedy within the existing library. 
    \item \textbf{Formulate and train.} The controller subagent receives development failure reports. By diagnosing the failure, it defines a reusable behavior, its switch conditions, and measurable success criteria. It sets up the training scene and initial robot states, and writes the behavior-specific rewards, curriculum, commands, and termination conditions. It then refines the training configuration using numerical measurements and visual evidence from controller rollouts within the acquisition budget (Appendix~\ref{app:acquisition_evidence}).
    \item \textbf{Evaluate and integrate.} The subagent evaluates checkpoints, writes controller integration code, and runs two confirmation sweeps over the development variants with an unchanged checkpoint, integration code, and confirmation program. The coordinator reviews its summary and README; promotion requires both sweeps to pass. If the controller is a prerequisite, acquisition of the dependent controller follows before task development resumes.
    \item \textbf{Resume or stop.} If acquisition succeeds, a fresh task subagent retries the task using the expanded library. Evidence from the previous attempt is archived separately, and the new attempt follows the same development, program selection, freezing, and held-out evaluation procedure. If the acquisition budget is exhausted or the declared criterion remains unreachable, the task is recorded as physically blocked; the controller is not reopened under another name. As the sequence proceeds to later tasks, acquired controllers and skills remain available for reuse.
\end{enumerate}

Figure~\ref{fig:app_workflow} in the main text summarizes the agent workflow and the refinement loops for task programs and controller training.

\subsection{Scope of Agent Autonomy}
\label{app:autonomy}

The framework supplies the workflow and controller-training templates, while the agents choose the behaviors to learn and design their training problems. Table~\ref{tab:app_constraints} summarizes the supplied constraints and the decisions left to the agents; detailed training and integration specifications appear in Appendix~\ref{app:controller_training}.

For example, on A02, the controller subagent decided to learn a wall-supported elevated stance (\ctrlrear{}) to raise the arm mount. It used the body-reconfiguration template, which requires a single transition followed by holding the resulting stance with the arm near home. Following this template, the subagent specified a posture with the front feet on the wall and hind feet on the floor, and designed rewards for elevation, wall contact, and stability.

\begin{table}[!ht]
    \centering
    \caption{Scope of agent autonomy within the supplied framework.}
    \label{tab:app_constraints}
    \footnotesize
    \setlength{\tabcolsep}{4pt}
    \begin{tabular}{@{}P{3.0cm}P{6.5cm}P{6.5cm}@{}}
        \toprule
        Component & Provided by framework & Agent decisions \\
        \midrule
        \textbf{Task program development} & Public task API; success criteria; development/held-out split; repair limits & Program logic, perception strategy, and controller commands; failure diagnosis and reusable skills \\
        \rowcolor{black!7}
        \textbf{Controller training} & Robot and simulator; joint-action interface; PPO settings; shared regularization and disturbances; frozen initial controller & Training launches, revisions, restarts, and resumptions; checkpoint selection \\
        {\leftskip=0.6em \emph{Body reconfiguration template}\par} & Uncommanded posture transition followed by holding; four-foot support; arm near home & Support structure and target posture; rewards, resets, terminations, assistance schedule, and success criteria \\
        \rowcolor{black!7}
        {\leftskip=0.6em \emph{Supported manipulation template}\par} & Entry from a settled supported stance; end-effector pose commands and arm-home mode; recorded-state resets & Supported workspace and target filtering; rewards, evaluation metrics, and termination conditions \\
        \textbf{Controller integration} & Controller registry; guarded switching mechanism; public observation interface; bounded target correction & Switch conditions; reusable Python functions; controller composition and confirmation programs \\
        \rowcolor{black!7}
        \textbf{Budgets and validation} & Training-step cap; development-only refinement; two successful confirmation sweeps before promotion & Allocation of training and debugging effort within the budget \\
        \bottomrule
    \end{tabular}
    \par\smallskip
    \begin{minipage}{\linewidth}
        \footnotesize
        Detailed numerical constraints and controller-training settings appear in Appendix~\ref{app:controller_training}.
    \end{minipage}
\end{table}

%% file: sessions/appendix/02_acquisition_and_reuse.tex
\clearpage
\section{Controller Acquisition and Reuse: Worked Examples}
\label{app:acquisition_evidence}

This section explains \method{} through controller acquisition and the development of two reusable skills in Claude Run~5. A02 follows the process from failed task programs to new controllers, the \texttt{wall\_high\_reach} skill, and renewed task execution. A04 adds a wall descent controller and the \texttt{floor\_recover} skill; B02 and B03 show how these acquired capabilities support later tasks. Training and integration specifications appear in Appendix~\ref{app:controller_training}; cross-run comparisons appear in Appendix~\ref{app:independent_runs}.

\subsection{Reading the Worked Examples}
\label{app:reading_examples}

The initial library contains the frozen \ctrlloco{} controller for standing locomotion and arm control, the supplied \texttt{floor\_pickup} skill, and perception and manipulation notes. A \emph{controller} is a learned policy that maps controller observations and supported commands to joint actions. A \emph{skill} is a reusable Python function that packages a task-level robot behavior using available controllers and environment feedback. Helper functions provide supporting perception, measurement, geometric, navigation, and pose-control operations for skills and task programs. Both skills and helpers can issue controller commands. A \emph{task program} is executable Python code for a particular task; it can invoke skills and include task-specific logic. Controller training adds physical behaviors; skills use them to accomplish subgoals such as reaching a target or picking up an object. The runtime checks switching guards, with one controller producing joint actions at a time.

The five steps below connect task failure to a reusable capability. All decisions and revisions use development feedback: controller evaluations assess the learned behavior, and confirmation sweeps test its integration on development variants. Held-out outcomes evaluate frozen task programs and do not guide development.

\subsection{A02: From Failed Reaching to a Reusable Skill}
\label{app:controller_acquisition}
\label{app:controller_acquisition_process}

\phantomsection
\label{app:gap_diagnosis}
\label{app:alternatives}
\noindent\textbf{Step 1: Diagnose the reaching failure.} A02 requires the end effector to reach a goal 1.25~m above the floor, within a 0.05~m position tolerance and with a horizontal approach toward the wall. Using only \ctrlloco{}, the initial task program left final position errors of 0.22--0.24~m, entirely vertical: the robot approached the wall and aligned the end effector, but could not reach the required height. Two revised programs also failed on all 15 development variants. The best revision reduced the error to 0.20--0.21~m, still about four times the tolerance. The persistent vertical shortfall across all three programs motivated a behavior that raises the arm mount.

\begin{figure}[!htbp]
    \centering
    \includegraphics[width=0.85\linewidth]{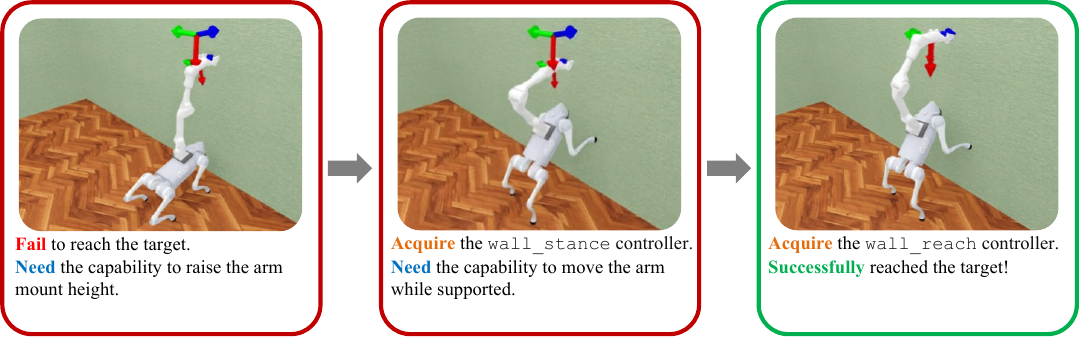}
    \caption{\textbf{A02, Steps 1--2: identify two complementary missing behaviors.} The initial controller falls short of the target. Wall-supported elevation (\ctrlrear{}) raises the arm mount; supported manipulation (\ctrlreach{}) then enables commanded arm motion from that stance.}
    \label{fig:app_capability_gap_a02}
\end{figure}

\noindent\textbf{Step 2: Formulate the new controllers.} The controller subagent divided the solution into two physical behaviors. The wall-supported elevation controller, \ctrlrear{}, places the front feet against the wall while keeping the hind feet on the floor, raising the arm mount with the arm homed. The supported manipulation controller, \ctrlreach{}, then tracks commanded end-effector poses while maintaining that stance. Elevation supplies the second controller's entry state; holding the elevated posture alone does not provide commanded manipulation (Fig.~\ref{fig:app_capability_gap_a02}).

Using the supplied templates, the agent formulated elevation rewards for pitch, shoulder height, support contacts, and stability, with temporary pose assistance during training. The supported manipulation controller is trained to track poses while retaining the supported body configuration, starting from recorded elevation states. The formulation, refinement, and validation process is applied first to wall-supported elevation and then to supported manipulation. Appendix~\ref{app:controller_configs} gives the final controller formulations.

\FloatBarrier
\clearpage

\begin{figure}[!htbp]
    \centering
    \includegraphics[width=0.85\linewidth]{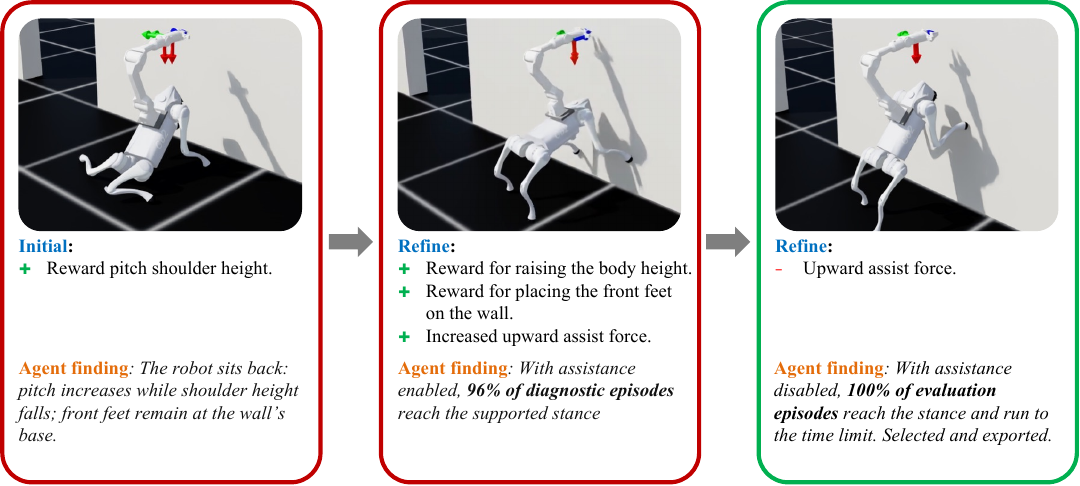}
    \caption{\textbf{A02, Step 3: sitting failure, supported stance, and assistance-free selection.} Left: pitch increases while shoulder height falls. Middle: after reward revisions, 96\% of diagnostic episodes reach the stance with assistance enabled. Right: with assistance disabled, 100\% of evaluation episodes reach it and run to the time limit. The last two panels follow one formulation revision and curriculum progression.}
    \label{fig:app_controller_progress}
\end{figure}

\begin{samepage}
\phantomsection
\label{app:a02_training_revisions}
\noindent\textbf{Step 3: Train, inspect, and revise.} Figure~\ref{fig:app_controller_progress} follows elevation training from left to right. Initially, the robot sat back on folded hind legs: pitch increased while shoulder height fell, and the front feet remained at the wall's base. The agent revised rewards for raising body height, measured at the shoulder, and placing the front feet on the wall, strengthened the upward assist force, and restarted training. With assistance enabled, 96\% of diagnostic episodes reached the supported stance (middle). The assistance curriculum then reduced the upward force to zero; the final controller passed unassisted evaluation and was selected and exported (right).
\par
\end{samepage}

\FloatBarrier

\noindent\textbf{Step 4: Validate the behaviors and integrate the controllers.} The controller subagent implemented the controller integration code and confirmation programs, with runtime guards checking each controller transition.

Each controller passed assistance-free evaluation and two confirmation sweeps over the same 15 development variants with an unchanged checkpoint, integration code, and confirmation program. Wall-supported elevation confirmation tested stance entry and maintenance with the arm homed; supported manipulation confirmation tested A02 completion. This validates the prerequisite before solving the original task (Appendix~\ref{app:checkpoint_evaluation}). The coordinator then added the validated controllers and their integration code to the library.

\begin{figure}[!htbp]
    \centering
    \includegraphics[width=0.85\linewidth]{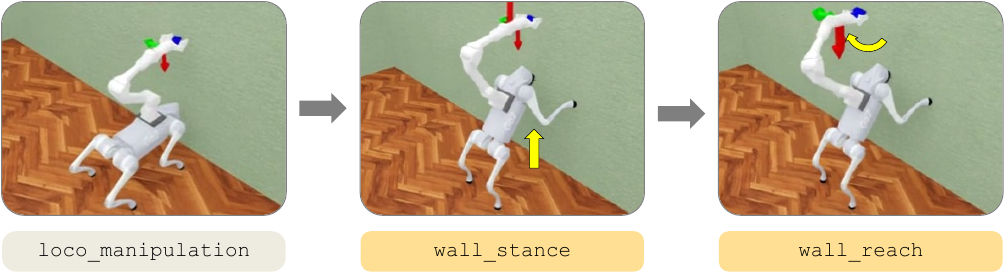}
    \caption{\textbf{A02, Step 5: develop the \texttt{wall\_high\_reach} skill during task-program refinement.} The skill aligns the robot with the wall using \ctrlloco{}, switches to \ctrlrear{} for elevation, and then switches to \ctrlreach{} to reach the target. The skill is reused for A03, B02, and B03.}
    \label{fig:app_skill_supported_reaching}
\end{figure}

\begin{samepage}
\phantomsection
\label{app:skill_composition}
\label{app:supported_reaching_skill}
\noindent\textbf{Step 5: Retry the task and retain the capability.} A fresh task subagent retried A02 with the expanded controller library and developed the \texttt{wall\_high\_reach} skill during task-program refinement. The skill first uses \ctrlloco{} to align the robot with the wall and home the arm, then switches to \ctrlrear{} to establish the supported stance, and finally switches to \ctrlreach{} to reach the target. Public observations guide skill execution, and runtime guards check each controller transition (Fig.~\ref{fig:app_skill_supported_reaching}). A02 improved from 0/50 held-out successes with the selected pre-acquisition program to 50/50 after acquisition. The coordinator retained the skill and its usage guidance for later tasks.
\par
\end{samepage}

\FloatBarrier
\clearpage
\subsection{A04: Acquiring Wall Descent}
\label{app:descent_example}

\begin{figure}[!htbp]
    \centering
    \includegraphics[width=0.85\linewidth]{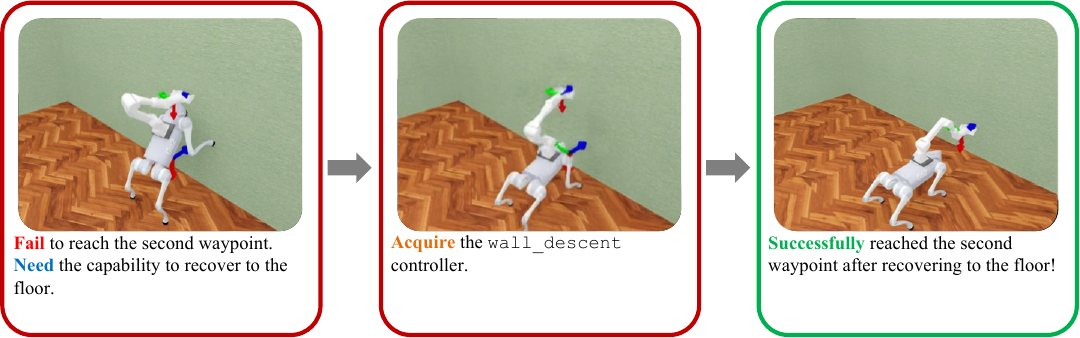}
    \caption{\textbf{A04: recover to the floor to reach the second waypoint.} Left: the supported stance cannot reach the second waypoint. Middle: wall descent (\ctrldismount{}) returns the robot to standing. Right: the initial controller can then reach the waypoint.}
    \label{fig:app_capability_gap_a04}
\end{figure}

\noindent\textbf{Diagnose the wall descent gap.} A04 requires an elevated waypoint followed by a waypoint 0.40~m above the floor. Reusing A02's \texttt{wall\_high\_reach} skill reached the first waypoint, but repeated program revisions failed at the second. That target was reachable while standing, yet runtime guards rejected direct switches from the supported stance to \ctrlloco{}. The missing capability was to recover to the floor: a wall descent controller must leave wall support before \ctrlloco{} can establish stable standing and reach the second waypoint (Fig.~\ref{fig:app_capability_gap_a04}).

\FloatBarrier

\begin{figure}[!htbp]
    \centering
    \includegraphics[width=0.85\linewidth]{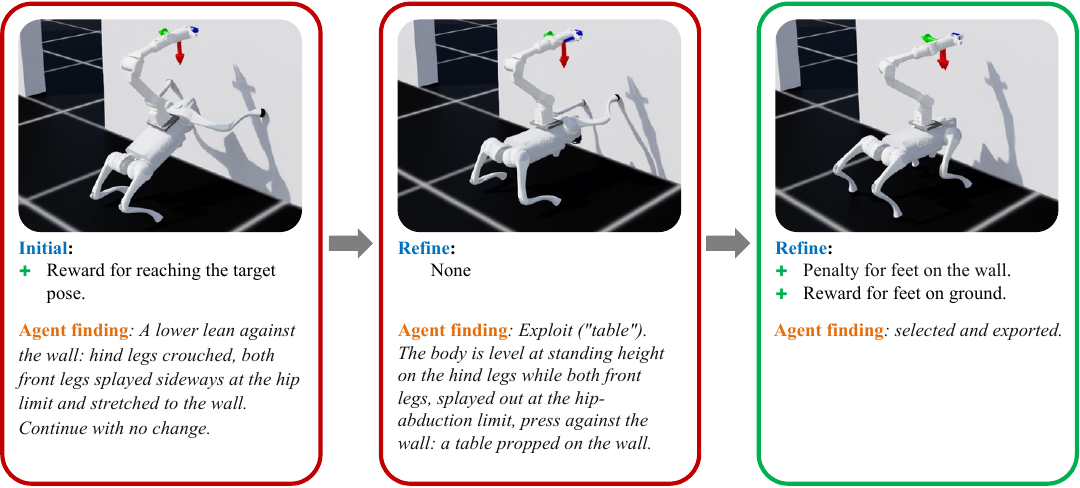}
    \caption{\textbf{A04: low lean, table exploit, and return to standing.} Left: the target-pose reward produces a low lean against the wall. Middle: continued training with no formulation change produces the wall-supported ``table'' posture. Right: contact-reward revisions penalize feet on the wall and reward feet on the ground. After further action-offset and posture refinements, the controller returns to a four-foot stance and is selected and exported (final configuration in Appendix~\ref{app:controller_configs}).}
    \label{fig:app_descent_revisions}
\end{figure}

\phantomsection
\label{app:descent_revisions}
\noindent\textbf{Acquire and refine wall descent.} The agent trained \ctrldismount{} from states recorded during rollouts of the supported manipulation controller with the arm homed, initially rewarding the target standing pose. Figure~\ref{fig:app_descent_revisions} follows three stages. Initially, the robot leaned low against the wall with crouched hind legs and front legs splayed sideways. The agent continued training without changing the formulation. The resulting reward exploit was a ``table'' posture: the body was level at standing height, but the front feet still pressed against the wall. The agent then penalized feet on the wall and rewarded feet on the ground. After further action-offset and posture refinements and retraining, the controller returned to a four-foot stance and was selected and exported (final configuration in Appendix~\ref{app:controller_configs}).

\FloatBarrier

\begin{figure}[!htbp]
    \centering
    \includegraphics[width=0.85\linewidth]{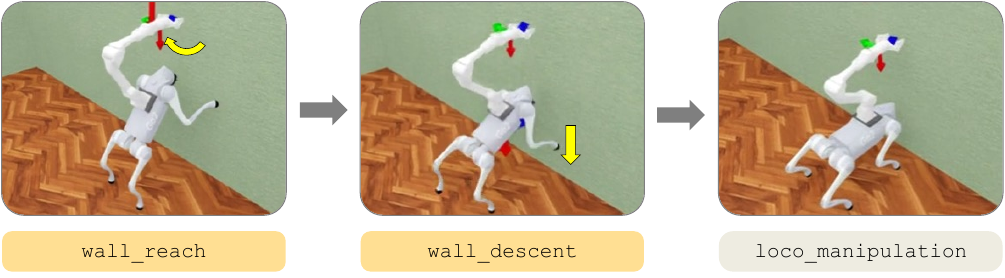}
    \caption{\textbf{The \texttt{floor\_recover} skill.} The skill moves the arm home using \ctrlreach{}, switches to \ctrldismount{} to descend, and then switches to \ctrlloco{} for the final force-stand action. The skill, acquired on A04, is reused unchanged after B02's elevated grasp.}
    \label{fig:app_skill_return_to_standing}
\end{figure}

\phantomsection
\label{app:return_to_standing_skill}
\noindent\textbf{Integrate the wall descent controller and retry A04.} After validation and development confirmation, the coordinator added the wall descent controller to the library. A fresh task subagent then retried A04 and developed the \texttt{floor\_recover} skill during task-program refinement. The skill first uses \ctrlreach{} to move the arm to its home pose, then switches to \ctrldismount{} to descend, and finally switches to \ctrlloco{} for a force-stand action that establishes a stable four-foot stance (Fig.~\ref{fig:app_skill_return_to_standing}). A04 can now compose \texttt{wall\_high\_reach} for the high waypoint with \texttt{floor\_recover} and subsequent reaching under \ctrlloco{} for the low waypoint.

\FloatBarrier

The resulting frozen task program improved A04 from 0/50 to 50/50 held-out successes. A04 adds the \ctrldismount{} controller and the \texttt{floor\_recover} skill with its integration code to the library while retaining A02's controllers and \texttt{wall\_high\_reach}. Its new lesson is that manipulating within a useful stance and leaving that stance require distinct physical capabilities.

\FloatBarrier
\subsection{Reusing the Accumulated Library}
\label{app:b_reuse}
\label{app:reuse_examples}
\label{app:capability_library}
\label{app:skill_growth}

The acquired controllers and skills support later tasks without further controller training. B02 and B03 reuse the library accumulated through A04, with controller weights and guards fixed. Task programs can still add perception, manipulation logic, and reusable helpers. Written guidance records applicability and failure signatures for later tasks.

\subsubsection{B02: Rack Retrieval with Existing Capabilities}

B02 reuses \texttt{wall\_high\_reach} to access the elevated object and \texttt{floor\_recover} to return to standing after grasping it. Their controller integration code is reused unchanged. New task logic adds wrist-camera localization, lateral centering, and grasp targets in a gravity-levelled frame. The measurement and centering routines were published as helpers for later tasks.

The \texttt{floor\_recover} skill uses the same homing, descent, and force-stand sequence developed on A04. The recovery skill originally developed for a low waypoint therefore also serves rack retrieval. B02 achieved 50/50 held-out successes.

\subsubsection{B03: Rack Placement with Reused Components}

B03 first picks up an object from the floor, using \texttt{floor\_pickup} routines repaired on B01, then reuses \texttt{wall\_high\_reach} and B02's measurement helpers. New logic models the held object and plans its placement. The program enters the stance beside the panel to avoid striking the rack board during elevation, places and releases the object, and ends in the supported stance. It therefore does not invoke \texttt{floor\_recover}. B03's placement procedure was documented, with no downstream reuse demonstrated.

B03 achieved 49/50 held-out successes. In the remaining variant, heading drift caused the elevation switch guard to reject the switch; the program ended with the robot still standing and holding the object.

The B02--B03 comparison shows how new tasks reuse acquired controllers, the \texttt{wall\_high\_reach} and \texttt{floor\_recover} skills, and selected helpers while supplying different task logic. Appendix~\ref{app:controller_switching} gives the controller-transition implementation.

\FloatBarrier

%% file: sessions/appendix/03_controller_training.tex
\clearpage
\section{Controller Training and Integration Specifications}
\label{app:controller_training}

\subsection{Policy Interface and Shared Optimization}
\label{app:training_settings}

All acquired controllers share the same 60-dimensional observation, comprising proprioceptive feedback and the previous action, and output 18-dimensional action vectors that the runtime converts to joint-position targets for the 12 Go2 leg joints and 6 PiPER arm joints. Commands differ by controller: body reconfiguration uses no commands, whereas supported manipulation receives a 10-dimensional command comprising one arm-motion flag and nine end-effector keypoint-error coordinates. Combining the shared observation with the controller-specific command yields 60 input dimensions for body reconfiguration and 70 for supported manipulation in Claude Run~5. The frozen \ctrlloco{} controller uses a 222-dimensional input. Tables~\ref{tab:app_policy} and~\ref{tab:app_ppo} give the recorded interfaces and supplied optimization settings. The agent can choose the default joint positions for each controller to help the robot transition between controllers.

\noindent\textbf{Pose-command encoding.} Three non-collinear keypoints encode both position and orientation. Their fixed offsets form an equilateral triangle in the end-effector's local $xy$-plane:
\[
    d_1=(r,0,0)^{\top},\quad
    d_2=(-r/2,\sqrt{3}r/2,0)^{\top},\quad
    d_3=(-r/2,-\sqrt{3}r/2,0)^{\top},\qquad r=0.20~\mathrm{m}.
\]
The commanded pose is supplied in the robot-root frame and transformed into the measured end-effector frame. Let $(p_t^{\mathrm{cmd}},R_t^{\mathrm{cmd}})$ denote its position and rotation matrix in that frame. The current keypoints are $d_i$, and the commanded keypoints are $p_t^{\mathrm{cmd}}+R_t^{\mathrm{cmd}}d_i$. Thus, the target-minus-current errors and full command are
\[
    e_{i,t}=p_t^{\mathrm{cmd}}+R_t^{\mathrm{cmd}}d_i-d_i,\qquad
    c_t=(m_t,e_{1,t}^{\top},e_{2,t}^{\top},e_{3,t}^{\top})^{\top},
\]
where $m_t$ is the scalar arm-motion flag selecting target tracking or arm homing. The nine error coordinates concatenate the $xyz$ error for each keypoint in the order above, replacing a separate position-and-quaternion command representation. When corruption of the controller command is enabled during training, the commanded keypoints receive a shared random translation of up to $\pm5$~mm per axis and a shared rotation of up to $\pm5^\circ$ in roll and pitch and $\pm2^\circ$ in yaw. The critic receives the uncorrupted command encoding, and this corruption is disabled for evaluation and task execution.

\begin{table}[!ht]
    \centering
    \caption{Policy interfaces and simulation settings.}
    \label{tab:app_policy}
    \footnotesize
    \setlength{\tabcolsep}{4pt}
    \begin{tabular}{@{}P{3.2cm}P{13.4cm}@{}}
        \toprule
        Item & Setting \\
        \midrule
        Action & Policy output: 18-dimensional vector $u_t$. Runtime joint-position targets: $q_t^{\mathrm{tgt}} = q^{\mathrm{default}} + 0.25\,u_t$, where $q^{\mathrm{default}}$ contains the controller-specific default joint positions \\
        \rowcolor{black!7}
        Rates & Physics 200~Hz, controllers 50~Hz, task programs 10~Hz \\
        Controller observation & Shared by all acquired controllers: base angular velocity (3), projected gravity (3), joint positions relative to default (18), joint velocities (18), previous action (18), totaling 60 dimensions \\
        \rowcolor{black!7}
        Controller command & Body reconfiguration: none. Supported manipulation: one arm-motion flag and three 3D target-minus-current keypoint errors in the measured end-effector frame ($1+9=10$ dimensions) \\
        Controller input & Shared observation plus controller-specific command: 60 dimensions for body reconfiguration and 70 for supported manipulation. The frozen \ctrlloco{} controller has its own 222-dimensional input \\
        \rowcolor{black!7}
        Critic input & Uncorrupted controller input plus privileged feedback: base linear velocity and foot contacts, plus controller-specific terms (e.g., base height, front-sensor clearance, foot distances to the wall, foot heights); 78--82 dimensions \\
        Parallel environments & 4,096 \\
        \rowcolor{black!7}
        Episode length & Chosen per controller \\
        \bottomrule
    \end{tabular}
\end{table}

\begin{table}[!ht]
    \centering
    \caption{PPO settings used for acquisition of all three controllers.}
    \label{tab:app_ppo}
    \footnotesize
    \setlength{\tabcolsep}{6pt}
    \begin{tabular}{@{}ll<{\hspace{\dimexpr2.5em-2\tabcolsep\relax}}ll@{}}
        \toprule
        Setting & Value & Setting & Value \\
        \midrule
        Rollout steps per environment & 24 & Clip parameter & 0.2 \\
        \rowcolor{black!7}
        Actor and critic MLP & 512-256-128, ELU & Value loss coefficient & 1.0 (clipped) \\
        Initial action std & 1.0 & Entropy coefficient & 0.01 \\
        \rowcolor{black!7}
        Learning rate & $10^{-3}$, adaptive & Learning epochs $\times$ minibatches & $5\times4$ \\
        Desired KL & 0.01 & Discount $\gamma$, GAE $\lambda$ & 0.99, 0.95 \\
        \rowcolor{black!7}
        Max gradient norm & 1.0 & Observation normalization & None \\
        Checkpoint interval & 100 iterations & Steps per iteration & 98,304 \\
        \bottomrule
    \end{tabular}
\end{table}

\begin{table}[!ht]
    \centering
    \caption{Shared domain randomization, disturbances, and fixed regularizers. BR: body reconfiguration; SM: supported manipulation. Regularizers without a category suffix apply to both.}
    \label{tab:app_randomization}
    \footnotesize
    \setlength{\tabcolsep}{4pt}
    \begin{tabular}{@{}P{5.1cm}P{2.7cm}@{\hspace{0.9cm}}P{5.1cm}P{2.7cm}@{}}
        \toprule
        \multicolumn{4}{@{}l}{\textbf{Domain randomization}} \\
        \midrule
        Parameter & Range & Parameter & Range \\
        \midrule
        Go2 static friction & $[0.6, 1.4]$ & PiPER static friction & $[0.5, 1.2]$ \\
        \rowcolor{black!7}
        Go2 dynamic friction & $[0.5, 1.2]$ & PiPER dynamic friction & $[0.5, 1.2]$ \\
        Go2 restitution & $[0, 0.15]$ & PiPER restitution & $[0, 0.1]$ \\
        \rowcolor{black!7}
        Go2 base mass & $+[-1, 3]$~kg & PiPER link mass & $\times[0.8, 1.2]$ \\
        Go2 base center of mass & $\pm0.05$~m per axis & PiPER link inertia & $\times[0.8, 1.2]$ \\
        \rowcolor{black!7}
        Go2 link inertia & $\times[0.8, 1.2]$ & PiPER end-effector payload & $+[0, 0.5]$~kg \\
        Go2 actuator stiffness & $\times[0.8, 1.2]$ & PiPER link center of mass & $\pm0.1$~m per axis \\
        \rowcolor{black!7}
        Go2 actuator damping & $\times[0.75, 1.5]$ & PiPER actuator stiffness & $\times[0.8, 1.2]$ \\
        Go2 joint friction & $[0, 0.05]$ & PiPER actuator damping & $\times[0.75, 1.5]$ \\
        \rowcolor{black!7}
        Push linear velocity ($x, y$) & Set to $\pm0.3$~m/s & PiPER joint friction & $[0, 0.05]$ \\
        Push yaw velocity & Set to $\pm0.3$~rad/s & & \\
        \midrule
        \multicolumn{4}{@{}l}{\textbf{Fixed regularizers (reward weights)}} \\
        \midrule
        Term & Weight & Term & Weight \\
        \midrule
        Support-polygon margin & $+0.8$ & Foot slide (BR) & $-0.1$ \\
        \rowcolor{black!7}
        Joint velocity & $-10^{-3}$ & Leg action rate (BR) & $-0.01$ \\
        Joint acceleration & $-2.5\cdot10^{-7}$ & Arm action rate (BR) & $-0.01$ \\
        \rowcolor{black!7}
        Joint torque & $-2.5\cdot10^{-4}$ & Leg posture prior (BR) & $-0.01$ \\
        Energy & $-2\cdot10^{-5}$ & Arm action rate (SM) & $-0.1$ \\
        \rowcolor{black!7}
        Joint limits & $-1.0$ & Initial leg posture (SM) & $+0.4$ \\
        Undesired contacts & $-1.0$ & End-effector excess linear speed (SM) & $-2.0$ \\
        \rowcolor{black!7}
        & & End-effector excess angular speed (SM) & $-0.05$ \\
        \bottomrule
    \end{tabular}
\end{table}

\subsection{Training Templates and Initialization Data}
\label{app:training_templates}

\noindent\textbf{Body reconfiguration.} This category learns one transition from its entry stance to a target stance, then holds that stance without commands or termination on success. The template limits orientation change to below 1.0~rad and requires at least 0.05~m front clearance to protect the front LiDAR and camera from contact with the support surface. The supplied starter rewards default base height, penalizes base tilt, and tracks the gravity-compensated arm-home pose through end-effector tracking. The agent replaces the behavior formulation with progress, contact, and stability terms for the chosen maneuver. The intended terminal contract requires all four feet in contact with the floor or supporting structure and no other supporting body contact. The success predicate also measures the pose change and must be false at reset.

\noindent\textbf{Supported manipulation.} This category tracks end-effector pose commands while preserving a supported stance. An arm-motion flag selects target tracking or return to the gravity-compensated home pose; the tracking reward remains active in both modes. With measured pose $(p,Q)$ and selected target $(p^\star,Q^\star)$, the weighted term is
\[
    r_{\mathrm{track}} = 3\exp\left[-\left(\frac{\|p-p^\star\|}{\sigma_p}\right)^2
    -\left(\frac{d_R(Q,Q^\star)}{\sigma_R}\right)^2\right],
\]
Here, $p$ and $p^\star$ are the measured and selected-target end-effector positions, $Q$ and $Q^\star$ are their orientation quaternions, and $d_R(Q,Q^\star)\in[0,\pi]$ is their angular orientation error in radians. The widths begin at $\sigma_p=0.40$~m and $\sigma_R=2.0$~rad and shrink by $0.7$ when normalized tracking reward exceeds 0.8, with floors of 0.10~m and 0.50~rad. These floors specify the curriculum limits, not a claim that every checkpoint reached them.

\noindent\textbf{Pose assistance helper.} We provide the agent with optional pose assistance, which it can choose to enable during training. A world-frame spring wrench can pull the base toward the target pose. When enabled, sustained behavior success reduces the stiffness until it reaches zero. Assisted diagnostic evaluations may be used to inspect intermediate training progress. Assistance is disabled for final controller validation, export, and task execution; if used during training, its curriculum must be completed before a checkpoint is eligible for validation and export.

\noindent\textbf{Reach-target data.} The supplied dataset contains 10,000 end-effector poses obtained by sampling fixed-base PiPER configurations within joint limits. Training perturbs sampled targets, clips lateral displacement to $|y|<0.35$~m, and rejects targets violating a 2~cm floor clearance or protection regions around the body and arm mount. The agent adds support-specific data and filtering: \ctrlreach{} adds 12,000 wall-facing targets verified by inverse kinematics in the reared stance, and rejects targets whose end-effector or fingertip point lies within 0.06 or 0.03~m of the wall.

\subsection{Final Controller Configurations}
\label{app:controller_configs}
\label{app:controllers}

The following list specifies the final behavior formulations. For \ctrlrear{}, stance progress rewards pitch only together with shoulder height (Appendix~\ref{app:a02_training_revisions}). For \ctrldismount{}, standing progress combines uprightness, base height, and the fraction of feet on the floor, while posture progress encourages the initial controller's standing leg configuration. Complete contact predicates, terminations, and reward functions are defined by the controller packages. The listed terminations supplement time-out, base contact, and the inherited 1.0~rad tilt limit; runtime guards are in Appendix~\ref{app:controller_switching}.

\begin{itemize}
    \item \ctrlrear{}
    \begin{itemize}
        \item \textbf{Motivating task:} A02 (prerequisite).
        \item \textbf{Behavior:} Rear against the wall: base pitched 0.68~rad nose-up (success 0.55--0.80~rad), shoulder $\geq 0.67$~m (target 0.73~m), front feet on the wall at least 0.25~m up, hind feet on the floor.
        \item \textbf{Rewards (weights):} Stance progress (pitch $\times$ shoulder height) 3.0; pitch 1.0; shoulder height 2.0; roll $-2.0$; front feet on wall 2.0; hind feet on floor 0.5; wall-foot slip $-0.2$; stance held 3.0; motion in the stance $-1.0$; wall standoff 0.5; sensor-clearance shortfall $-2.0$; arm home terms 3.0.
        \item \textbf{Curriculum:} Pose assist on the base (initial stiffness 800~N/m $z$, 150~N/m $x$, 80 / 30~N\,m/rad pitch / roll), $\times0.8$ when success exceeds 0.6, down to zero.
        \item \textbf{Terminations:} Contact of any body other than the feet and gripper fingers above 1~N; sensor clearance below 0.05~m; base below 0.10~m.
    \end{itemize}
    \item \ctrlreach{}
    \begin{itemize}
        \item \textbf{Motivating task:} A02 (closes the gap).
        \item \textbf{Behavior:} Track commanded end-effector poses while holding the stance; fold the arm home when arm motion is off.
        \item \textbf{Rewards (weights):} End-effector tracking 3.0; arm home joint $-0.01$; alive 0.05; front feet on wall 0.5; hind feet on floor 0.5; tilt above 0.74~rad $-5.0$.
        \item \textbf{Curriculum:} Tracking widths 0.40 $\rightarrow$ 0.10~m and 2.0 $\rightarrow$ 0.50~rad.
        \item \textbf{Terminations:} Base drift above 0.15~m or 0.35~rad; contact of any body other than the feet and gripper fingers above 1~N.
    \end{itemize}
    \item \ctrldismount{}
    \begin{itemize}
        \item \textbf{Motivating task:} A04 (closes the gap).
        \item \textbf{Behavior:} Come down to standing: tilt at most 0.15~rad, base 0.22--0.34~m, four feet on the floor, base $\geq 0.43$~m from the wall.
        \item \textbf{Rewards (weights):} Standing progress (tilt $\times$ height $\times$ feet on floor) 3.0; upright 1.0; wall standoff 1.0; posture progress 2.0; feet on floor 2.0; feet on wall $-1.0$; standing held 3.0; motion while standing $-1.0$; standing posture 2.0; sensor-clearance shortfall $-2.0$; arm home terms 3.0.
        \item \textbf{Curriculum:} None (the motion lowers the center of mass).
        \item \textbf{Terminations:} Contact of any body other than the feet and gripper fingers above 1~N; sensor clearance below 0.05~m; base below 0.10~m.
    \end{itemize}
\end{itemize}

\subsection{Checkpoint Evaluation and Acceptance Criteria}
\label{app:checkpoint_evaluation}

Before training, the controller subagent declares a controller-level evaluation criterion and a development-sweep criterion. A prerequisite controller must demonstrate its intended capability; the controller that closes the gap must enable full task success. Promotion requires two confirmation sweeps over the same 15 development variants with one unchanged checkpoint, integration code, and confirmation program. These are repeated development checks, not independent held-out generalization tests. The list below reports the Claude Run~5 controller-level measurements and development confirmation outcomes; final task outcomes appear in Appendix~\ref{app:acquisition_evidence}.

The agent performs configuration and simulation checks, a short PPO pilot, longer training, and checkpoint evaluation. Every tenth saved checkpoint and the final checkpoint receive evaluation with keyframes, and the agent may evaluate earlier checkpoints for diagnosis; numerical returns alone are insufficient for a verdict. The final controller evaluations reported below use 64 parallel environments and seed 42, stopping after at least 100 completed episodes. Because the evaluator counts all completions in each vectorized step, this threshold can be exceeded; each reported final evaluation contains 128 episodes. Assistance is disabled for these final evaluations; assisted diagnostic evaluations during training, such as the intermediate result in Appendix~\ref{app:a02_training_revisions}, do not establish controller acceptance. Controller-level disturbances are distinct from the nominal-physics task evaluations described in Appendix~\ref{app:task_definitions}.

Each controller has a total budget of $1.5\times10^9$ environment steps, summed across pilots, restarts, and training segments; evaluation does not count toward that training budget. A standard iteration uses 4,096 environments and 24 rollout steps per environment, or 98,304 environment steps. Training ran on one RTX 5090 with 32~GB memory, taking approximately 1.3--1.6~s per standard iteration across Runs~1--5. Jobs are otherwise exclusive, but one checkpoint evaluation may share the GPU with its active training run. In Runs~4 and~5, controller subagents also ran task replays, such as routing checks and certification replays, while their training runs were active. Actual acquisition effort is reported in Appendix~\ref{app:compute}.

\phantomsection
\label{app:criteria}
The wall-supported elevation controller was confirmed on its prerequisite capability with the arm homed; supported manipulation and wall descent were confirmed by A02 and A04 task success, respectively.

\begin{itemize}
    \item \ctrlrear{}
    \begin{itemize}
        \item \textbf{Declared evaluation criterion:} At least 95\% of episodes end only by time-out; instantaneous stance success at least 0.75 over all steps.
        \item \textbf{Measured evaluation:} 128/128 time-outs; stance success 0.979; every episode reached the stance, after 0.26~s on average.
        \item \textbf{Required development confirmation:} Stance within 3~s of the switch, held to the episode end; return to \ctrlloco{} rejected.
        \item \textbf{Observed development confirmation:} 15/15 twice; stance within 0.7~s, held 11.6--13.6~s; return rejected.
    \end{itemize}
    \item \ctrlreach{}
    \begin{itemize}
        \item \textbf{Declared evaluation criterion:} At least 90\% of episodes end only by time-out.
        \item \textbf{Measured evaluation:} 128/128 time-outs; A02-like targets within tolerance on 93.8\% of arm-motion steps, mean error 0.024~m and 0.086~rad.
        \item \textbf{Required development confirmation:} A02 success after a guarded switch from \ctrlrear{}.
        \item \textbf{Observed development confirmation:} 15/15 twice; goal completed at 8.2--10.9~s.
    \end{itemize}
    \item \ctrldismount{}
    \begin{itemize}
        \item \textbf{Declared evaluation criterion:} At least 95\% of episodes end only by time-out; at least 95\% reach standing; mean time to stand $\leq2.0$~s.
        \item \textbf{Measured evaluation:} 128/128 time-outs; every episode reached and ended standing; mean time to stand 0.23~s.
        \item \textbf{Required development confirmation:} A04 success, with descent entry and hand-back accepted after only guard rejections.
        \item \textbf{Observed development confirmation:} 15/15 twice; \ctrlloco{} selected 0.8--1.0~s after descent began.
    \end{itemize}
\end{itemize}

For the elevation sweeps, the required stance had nose-up pitch in $[0.55,0.80]$~rad, absolute roll at most 0.20~rad, shoulder height at least 0.67~m, and front-sensor clearance at least 0.05~m.

\FloatBarrier
\subsection{Controller Switching}
\label{app:controller_switching}

The robot can switch to another controller only after satisfying all of that controller's switch conditions. The controller subagent defines and implements these conditions, which the runtime checks before allowing a switch. Exactly one controller produces joint actions at a time; rejected requests retain the active controller and are recorded in the trace.

For elevated reaching, \texttt{wall\_high\_reach} aligns the robot using \ctrlloco{}, establishes wall support using \ctrlrear{}, and switches to \ctrlreach{} for arm motion. For recovery, \texttt{floor\_recover} homes the arm using \ctrlreach{}, switches to \ctrldismount{} to descend, and returns control to \ctrlloco{} for the final force-stand action.

The runtime enforces the following switch conditions and hold durations:\par
\begingroup
\renewcommand{\iedlistdecl}{%
    \setlength{\topsep}{0pt}%
    \setlength{\itemsep}{0pt}%
    \setlength{\parsep}{0pt}%
    \setlength{\partopsep}{0pt}%
}
\noindent
\begin{minipage}[t]{0.48\linewidth}
\vspace{0pt}
\begin{itemize}
    \setlength{\itemsep}{4pt}
    \item \ctrlloco{} $\rightarrow$ \ctrlrear{}
    \begin{itemize}
        \item Front-sensor distance 0.12--0.20~m.
        \item Front-sensor $|$angle$| \leq 0.10$~rad.
        \item Base stopped.
        \item Arm homed.
        \item Hold for 1.0~s.
    \end{itemize}
    \item \ctrlrear{} $\rightarrow$ \ctrlreach{}
    \begin{itemize}
        \item Joint speeds below 1.5~rad/s.
        \item Hold for 0.5~s.
    \end{itemize}
\end{itemize}
\end{minipage}\hfill
\begin{minipage}[t]{0.48\linewidth}
\vspace{0pt}
\begin{itemize}
    \setlength{\itemsep}{4pt}
    \item \ctrlreach{} $\rightarrow$ \ctrldismount{}
    \begin{itemize}
        \item Arm motion off.
        \item Arm homed.
        \item Stance held.
        \item Joint speeds below 1.5~rad/s.
        \item Hold for 0.5~s.
    \end{itemize}
    \item \ctrldismount{} $\rightarrow$ \ctrlloco{}
    \begin{itemize}
        \item Tilt $\leq 0.20$~rad.
        \item Every joint speed below 1.0~rad/s.
        \item Hold for 0.3~s.
    \end{itemize}
\end{itemize}
\end{minipage}
\par
\endgroup

%% file: sessions/appendix/04_experimental_protocols.tex
\clearpage
\section{Experimental Protocols}
\label{app:evaluation_deployment}

This section specifies the simulation tasks, evaluation procedures, and baseline results. Appendix~\ref{app:run_study} reports acquisition outcomes, artifact inventories, revisions, training effort, runtime, and model-usage estimates across five independent Claude runs. Claude Run~5 provides the detailed acquisition and reuse narrative in Appendix~\ref{app:acquisition_evidence}; training specifications appear in Appendix~\ref{app:controller_training}.

\subsection{Task Definitions}
\label{app:task_definitions}

All tasks share the same room and Go2--PiPER platform. The manipulation objects have body diameter 0.065~m and mass 0.5~kg; the bottle is 0.200~m tall and the cans 0.113--0.121~m tall. The B02/B03 rack has a $1\times0.12$~m board with its top 1.05~m above the floor, mounted on a $1\times1.5$~m panel flush with the wall. Table~\ref{tab:app_tasks} gives the task-instance distributions and success rules. For A01--A04, position and orientation tolerances are 0.05~m and 0.20~rad.

\begin{table}[!ht]
    \centering
    \caption{Simulation task variants. Every success condition must hold continuously for 1~s; success then latches.}
    \label{tab:app_tasks}
    \footnotesize
    \setlength{\tabcolsep}{4pt}
    \begin{tabular}{@{}P{1.0cm}P{2.8cm}P{5.2cm}P{0.9cm}P{6.0cm}@{}}
        \toprule
        Task & Start & Varied per variant & Limit & Success; failure \\
        \midrule
        A01 & $x = -1$~m, facing $+x$ & Target $x \in [0, 2]$, $y \in [-1, 1]$, $z \in [0.3, 0.6]$~m; yaw $\in [0, 2\pi]$ applied to the A02 orientation & 20~s & End effector within 0.05~m and 0.20~rad of the target \\
        \rowcolor{black!7}
        A02 & $x, y \in [-0.5, 0.5]$~m, yaw $\in [-36^\circ, 36^\circ]$ & Start pose & 20~s & Fixed goal 0.30~m from the wall and 1.25~m above the floor, level gripper pointing at the wall \\
        A03 & As A02 & Start pose; waypoints 2 and 3 offset from the A02 goal by $\Delta x, \Delta y \in [-0.15, 0.15]$~m and $\Delta z \in [-0.30, -0.15]$~m & 20~s & Three waypoints in order, each within tolerance for 1~s \\
        \rowcolor{black!7}
        A04 & As A02 & Start pose & 20~s & Waypoints at 1.25~m then 0.40~m above the floor, same wall column and orientation \\
        B01 & $x = 0.5$~m, facing $+x$ & Object position $\pm0.2$~m around $(1.5, 0)$~m; object (bottle with probability 0.5, else one of three cans) & 40~s & Held object's lowest point at least 0.30~m above the floor; failure if the unheld object tips past $45^\circ$ \\
        \rowcolor{black!7}
        B02 & $x = 1.6$~m, facing $+x$ & Object position $\pm0.3$~m along the board; object type & 60~s & Object held, base stopped and arm homed, four feet on the floor, base upright; failure if the unheld object tips past $45^\circ$ or falls below 1.0~m \\
        B03 & $x = 1.35$~m, facing $+x$ & Object position $\pm0.01$~m ($x$), $\pm0.2$~m ($y$) around $(2.35, 0)$~m; object type & 60~s & Released object upright within $10^\circ$ with its base within 2~cm of the board top; failure if the unheld object drops to the floor after the first pickup \\
        \bottomrule
    \end{tabular}
\end{table}

For B03, the released-object height and uprightness conditions do not include a horizontal-position or board-footprint containment test. Evaluation ends when the 1~s success dwell completes; it does not establish longer-term placement stability or require the robot to return to standing.

\phantomsection
\noindent\textbf{Simulation assumptions.}
\label{app:sim_assumptions}
Physics randomization, push disturbances, curricula, and additive observation noise are disabled during task execution. Object masks are simulator ground-truth segmentation; B01--B03 do not expose ground-truth object or rack poses. Controller-training randomization, disturbances, and assistance are specified separately in Appendix~\ref{app:controller_training}.

\subsection{Evaluation and Baseline Protocols}
\label{app:evaluation_protocol}

A task name and seed determine the settled start state and task inputs. Seeds 51--65 define the 15 development variants; seeds 1--50 define the 50 held-out variants. Resets preserve a variant across retries.

\noindent\textbf{Development and selection.} Stage~0 captures a scene snapshot on variant 51, records an initial analysis, and executes an initial program on all development variants. Stage~1 allows at most three repair replays per failed variant. Exhausted failures receive a physical, perceptual, or algorithmic classification. One task program is selected by most development successes, then fewest crashes, then simplest observation-driven behavior. There is no separate confirmation sweep for task programs; the paired sweeps in Appendix~\ref{app:checkpoint_evaluation} apply to controller acquisition.

\noindent\textbf{Held-out evaluation.} After development, the coordinator publishes reusable findings and runs the frozen program once per held-out variant. Each replay records its program and library snapshot; an immutable manifest identifies the held-out run by program, library, task, and seed-list hashes. Completion means that all 50 results exist, regardless of success count. A controller promotion leads to archived earlier evidence and a fresh task-development attempt. The reported final result is the last frozen program's held-out run; earlier evaluations with the pre-acquisition controller library remain evidence of the preceding capability set. Neither those outcomes nor the final outcomes are used to select repairs or justify acquisition.

\phantomsection
\noindent\textbf{Baseline comparison.}
\label{app:baseline_protocol}
We implement ASPIRE and Code-as-Policies as controlled variants of \method{} within the same agent and evaluation framework. All three use Claude Opus~5.5 (\texttt{claude-opus-5-5}) at reasoning effort \texttt{xhigh} through Claude Code, the same coordinator and fresh-task-agent structure, and common task-program instructions, except for the acquisition and repair permissions described below. Each starts with the frozen \ctrlloco{} controller, the supplied \texttt{floor\_pickup} skill, and the perception and manipulation notes described in Appendix~\ref{app:initial_instructions}. All follow A01 $\rightarrow$ A02 $\rightarrow$ A03 $\rightarrow$ A04 $\rightarrow$ B01 $\rightarrow$ B02 $\rightarrow$ B03 and retain reusable Python skills within a run; acquired artifacts are not shared between independent runs.

ASPIRE differs from \method{} only by disabling controller acquisition and the resulting controller-library expansion. It uses the same initial program synthesis, development-feedback repair, program selection, and reusable-skill accumulation. Stage~1 permits at most three repair replays per failed development variant, with no additional evolutionary-search stage. The initial low-level controller remains frozen throughout the task sequence. For \method{}, controller acquisition follows Appendix~\ref{app:checkpoint_evaluation}, and successful acquisition permits a fresh task-development attempt with the expanded library.

Code-as-Policies uses the ASPIRE configuration with feedback-based program repair additionally disabled. It generates one initial program per task, rather than one per variant, and executes that program unchanged on all 15 development variants and all 50 held-out variants. It permits no repair replays, feedback-driven candidate selection, or revisions to the library code invoked by the current task. The program can still react to observations during execution, and reusable findings can be retained for later tasks under the same skill-accumulation procedure.

All three configurations share the public task API and observations (Appendix~\ref{app:runtime_interfaces}), task definitions and success rules (Table~\ref{tab:app_tasks}), simulation assumptions, settled start states for each seed, development seeds 51--65, and held-out seeds 1--50. Episode limits are 20~s for A01--A04, 40~s for B01, and 60~s for B02--B03. ASPIRE and \method{} use the development-selection rule above; Code-as-Policies freezes its sole initial program. Each frozen program is evaluated once per held-out variant, without using held-out outcomes for repair, skill updates, or acquisition. These settings match task evaluation, while the repair allowance differs for Code-as-Policies and controller training and subsequent task development add computation for \method{}; total computation is not matched. The shared starting conditions across the five \method{} runs are detailed in Appendix~\ref{app:run_conditions}.

\noindent\textbf{Baseline run results.} Table~\ref{tab:app_baseline_runs} reports five completed independent runs for each baseline. Each entry gives the final frozen program's held-out successes. Table~\ref{tab:task_success} summarizes these counts and the five Claude \method{} runs as mean success rates and sample standard deviations across runs. Overall rates are computed per run from all 350 task trials before calculating the mean and standard deviation.

\begin{table}[!ht]
    \centering
    \caption{\textbf{Held-out baseline successes by run.} Each task uses 50 variants, giving 350 trials per run and 1,750 trials per baseline across five runs. Entries are success counts.}
    \label{tab:app_baseline_runs}
    \footnotesize
    \setlength{\tabcolsep}{7pt}
    \begin{tabular}{@{}ll*{7}{r}r@{}}
        \toprule
        Method & Run & A01 & A02 & A03 & A04 & B01 & B02 & B03 & Total \\
        \midrule
        Code-as-Policies & 1 & 42 & 0 & 0 & 0 & 50 & 0 & 0 & 92 \\
                       & 2 & 35 & 0 & 0 & 0 & 48 & 0 & 0 & 83 \\
                       & 3 & 36 & 0 & 0 & 0 & 28 & 0 & 0 & 64 \\
                       & 4 & 28 & 0 & 0 & 0 & 45 & 0 & 0 & 73 \\
                       & 5 & 42 & 0 & 0 & 0 & 49 & 0 & 0 & 91 \\
        \midrule
        ASPIRE         & 1 & 50 & 0 & 0 & 0 & 50 & 0 & 0 & 100 \\
                       & 2 & 50 & 0 & 0 & 0 & 50 & 0 & 0 & 100 \\
                       & 3 & 49 & 0 & 0 & 0 & 50 & 0 & 0 & 99 \\
                       & 4 & 50 & 0 & 0 & 0 & 48 & 0 & 0 & 98 \\
                       & 5 & 50 & 0 & 0 & 0 & 50 & 0 & 0 & 100 \\
        \bottomrule
    \end{tabular}
\end{table}

%% file: sessions/appendix/05_independent_runs.tex
\clearpage
\section{Independent Acquisition Runs}
\label{app:run_study}
\label{app:independent_runs}

This section reports five independent controller-acquisition runs using Claude Opus 5.5 through Claude Code. All five runs completed A01--B03 and form the main-text aggregates. Claude Run~5 supplies the worked examples in Appendix~\ref{app:acquisition_evidence}.

\providecommand{\appartifact}[1]{\nolinkurl{#1}}

\subsection{Study Design and Run Conditions}
\label{app:run_conditions}

\noindent\textbf{Independent acquisition runs.} Each run starts from a fresh repository without controllers, reusable modules, or findings acquired by another run. Runs share task definitions, success evaluators, development variants 51--65, held-out variants 1--50, controller-training budgets, and promotion rules. A run includes the complete A01--B03 sequence and retains its own acquired library across tasks. Each task contributes 50 final held-out trials, giving 350 trials per run and 1,750 trials across the five runs. The acquisition run is the unit of replication; task variants and controller-confirmation sweeps do not increase the number of independent acquisitions. All controller-training launches, including pilots, and checkpoint evaluations use seed~42.

\noindent\textbf{Claude run conditions.} All five use \texttt{claude-opus-5-5} at reasoning effort \texttt{xhigh} and share the same initial repository configuration, including the supplied training framework, initial datasets, frozen loco-manipulation controller checkpoint, pickup skill, and public task interfaces. The supplied pickup skill uses a minimum walking command of 0.08~m/s, and front-surface distance and heading are available to B-task programs in every run. Task success rules, held-out seeds, and training budgets are identical across runs.

\subsection{Task Performance and Repeatability}
\label{app:run_performance}

Table~\ref{tab:app_run_comparison} preserves every run's final frozen-program outcomes, including the pre-acquisition results for A02 and A04. Every completed run acquired wall-supported elevation, supported manipulation, and wall descent without exhausting a controller's budget. None trained an additional controller for A03 or B01--B03. The remaining task failures occurred in completed executions.

\begin{table}[!ht]
    \centering
    \caption{Held-out successes by acquisition run. Task entries are counts out of 50; arrows preserve before$\rightarrow$after acquisition outcomes. Overall rates use the final seven programs (350 trials).}
    \label{tab:app_run_comparison}
    \footnotesize
    \setlength{\tabcolsep}{5pt}
    \begin{tabular}{@{}l*{7}{c}rr@{}}
        \toprule
        Run & A01 & A02 & A03 & A04 & B01 & B02 & B03 & Total & Rate (\%) \\
        \midrule
        1 & 50 & $0\rightarrow50$ & 47 & $0\rightarrow50$ & 46 & 49 & 50 & 342/350 & 97.71 \\
        2 & 49 & $0\rightarrow50$ & 50 & $0\rightarrow50$ & 50 & 50 & 49 & 348/350 & 99.43 \\
        3 & 50 & $0\rightarrow50$ & 49 & $0\rightarrow50$ & 50 & 50 & 49 & 348/350 & 99.43 \\
        4 & 50 & $0\rightarrow50$ & 50 & $0\rightarrow50$ & 48 & 50 & 49 & 347/350 & 99.14 \\
        5 & 49 & $0\rightarrow50$ & 50 & $0\rightarrow50$ & 50 & 50 & 49 & 348/350 & 99.43 \\
        \bottomrule
    \end{tabular}
\end{table}

\begin{table}[!ht]
    \centering
    \caption{Held-out success rates (\%) as means $\pm$ sample standard deviations across the five independent acquisition runs.}
    \label{tab:app_run_summary}
    \footnotesize
    \setlength{\tabcolsep}{12pt}
    \begin{tabular}{@{}lr@{}}
        \toprule
        Task & Success rate ($n=5$) \\
        \midrule
        A01 & $99.2\pm1.1$ \\
        A02 & $100.0\pm0.0$ \\
        A03 & $98.4\pm2.6$ \\
        A04 & $100.0\pm0.0$ \\
        B01 & $97.6\pm3.6$ \\
        B02 & $99.6\pm0.9$ \\
        B03 & $98.4\pm0.9$ \\
        \midrule
        Overall & $99.0\pm0.7$ \\
        \bottomrule
    \end{tabular}
\end{table}

Task-level summaries use one success rate per acquisition run; overall summaries first divide each run's final successes by 350. Sample standard deviations use the run-level denominator $n-1$, with $n=5$, and summarize variability across independent acquisition runs under the same starting configuration.

\subsection{Acquired Controllers, Skills, and Helpers}
\label{app:run_artifacts}

Table~\ref{tab:app_capability_inventory} maps exact controller identifiers to shared functional roles. All five runs acquired wall-supported elevation and supported manipulation during A02, then wall descent during A04. Each of their three controllers passed two confirmation sweeps over the same 15 development variants with one unchanged checkpoint, integration code, and confirmation program. Elevation sweeps confirm the prerequisite stance with the arm homed; supported manipulation and wall descent sweeps confirm composed A02 and A04 completion. These are development checks, separate from held-out task performance. All five runs reused acquired controllers on B02 without further controller training.

\begin{table}[!ht]
    \centering
    \caption{Exact acquired-controller identifiers by capability role. Every run adds three controllers. Pitch is the documented nose-up elevation target magnitude, not a common performance metric; runtime sign conventions differ.}
    \label{tab:app_capability_inventory}
    \footnotesize
    \setlength{\tabcolsep}{4pt}
    \begin{tabular}{@{}lP{4.0cm}P{4.2cm}P{4.1cm}r@{}}
        \toprule
        Run & Wall-supported elevation (A02) & Supported manipulation (A02) & Wall descent (A04) & Pitch (rad) \\
        \midrule
        Claude~1 & \appartifact{wall_rear_up} & \appartifact{wall_rear_reach} & \appartifact{wall_dismount} & 0.70 \\
        Claude~2 & \appartifact{wall_rear_stance} & \appartifact{wall_rear_reach} & \appartifact{wall_rear_stand_down} & 0.85 \\
        Claude~3 & \appartifact{wall_rear_stance} & \appartifact{wall_rear_reach} & \appartifact{wall_rear_dismount} & 0.85 \\
        Claude~4 & \appartifact{wall_rear_up} & \appartifact{wall_stance_reach} & \appartifact{wall_step_down} & 0.70 \\
        Claude~5 & \appartifact{wall_stance} & \appartifact{wall_reach} & \appartifact{wall_descent} & 0.68 \\
        \bottomrule
    \end{tabular}
\end{table}

\noindent\textbf{Introduced skills and helpers.} The following list describes the reusable skills and helpers introduced in the five runs. Skills package task-level robot behaviors; helpers provide supporting perception, geometric, navigation, or pose-control operations. Both can issue controller commands. Claude Run~5's two acquired skills, \texttt{wall\_high\_reach} and \texttt{floor\_recover}, are described in detail in Appendix~\ref{app:acquisition_evidence}.

\begin{itemize}
    \item \textbf{Claude Run~1}
    \begin{samepage}
    \begin{itemize}
        \item \appartifact{reach_stance}: Reach a target with 4 legs standing. A skill introduced in A01 and reused in A04. 
        \item \appartifact{level_grasp}: Keeps the gripper level from a tilted stance and tracks the finger centre during object placement. A helper function introduced in B03.
        \item \appartifact{upright_place}: Places an object on the rack from a tilted stance, using stance measurements to determine the rack board height. A skill introduced in B03.
    \end{itemize}
    \end{samepage}
    \item \textbf{Claude Run~2}
    \begin{samepage}
    \begin{itemize}
        \item \appartifact{base_motion}: Provides wall-approach motion and base motion after returning to standing. A helper function used in A02--A04.
        \item \appartifact{eef_pose}: Provides end-effector pose-control routines and rotation calculations. A helper function whose rotation utilities are reused in A02--B03.
        \item \appartifact{held_object}: Records grasp geometry at gripper closure during floor pickup and uses it to plan placement from a tilted stance. A helper function introduced in B03. 
    \end{itemize}
    \end{samepage}
    \item \textbf{Claude Run~3}
    \begin{samepage}
    \begin{itemize}
        \item \appartifact{base_nav}: Provides base navigation and arm holding while standing. A helper function used in A01 and A04 for tracking the target near the floor.
        \item \appartifact{shelf_reach}: Provides rack-grasp routines from a wall-supported stance, with the hind feet on the floor and front feet against the wall. A skill introduced in B02. B03 reuses its measurements during floor pickup and placement from a tilted stance.
        \item \appartifact{wall_shelf_place}: Composes floor pickup, elevation, reaching, and rack placement. A skill introduced in B03.
    \end{itemize}
    \end{samepage}
    \item \textbf{Claude Run~4}
    \begin{samepage}
    \begin{itemize}
        \item \appartifact{base_stance}: Provides base positioning before elevation. A helper function used in A02--A04, B02, B03.
        \item \appartifact{eef_hold}: Provides standing reach and end-effector holding. A helper function used in A01 and A04 for tracking low targets.
        \item \appartifact{level_reach}: Provides reaching for rack retrieval and placement from a tilted stance. A skill first introduced in B02 and B03 reuses it for placement from a tilted stance.
    \end{itemize}
    \end{samepage}
    \item \textbf{Claude Run~5}
    \begin{samepage}
    \begin{itemize}
        \item \appartifact{base_nav}: Navigates to a wall-facing stance and walks after returning to standing. A helper function used in \appartifact{wall_high_reach}
        \item \appartifact{eef_reach}: Provides end-effector pose-control routines for high goals from a wall-supported stance and low goals with all feet on the floor. A helper function used in \appartifact{wall_high_reach}.
        \item \appartifact{wall_high_reach}: Composes wall alignment, supported elevation, and reaching to commanded targets. Developed in A02, it is reused for the elevated waypoints in A03 and A04, retrieval in B02, and placement in B03.
        \item \appartifact{floor_recover}: Homes the arm, descends from wall support, and restores a stable four-foot stance. Developed in A04 to enable the subsequent low waypoint, it is reused unchanged after B02's elevated grasp while retaining the object.
    \end{itemize}
    \end{samepage}
\end{itemize}

The supplied \appartifact{floor_pickup} skill was also revised in Claude Runs~2, 3, and~5.

\begin{figure}[!htbp]
    \centering
    \includegraphics[width=\linewidth]{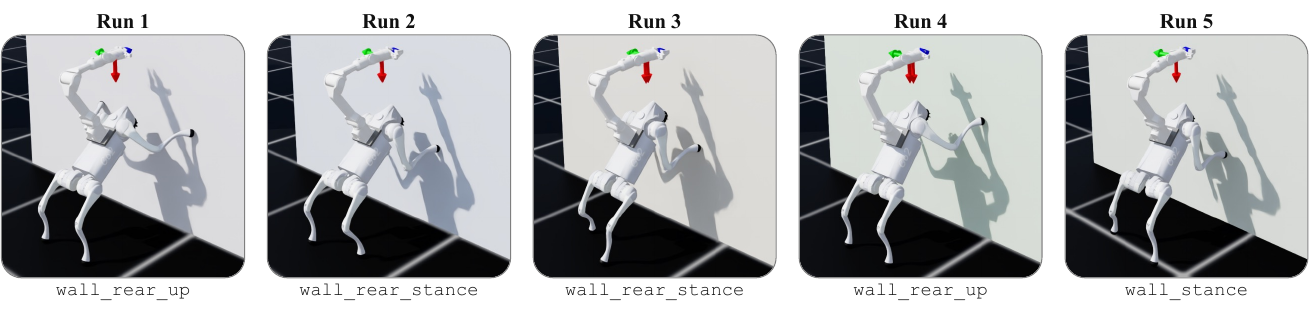}
    \caption{\textbf{Independent acquisition of wall-supported elevation.} Panels show the five runs' exported-controller confirmations on A02 development variant~51 with training assistance disabled and the arm homed. Final-frame times for Runs~1--5 are 19.7, 19.6, 20.0, 19.5, and 19.7~s, respectively. Every shown controller passed two unchanged 15-variant elevation sweeps. Cameras follow the base; framing is comparable but viewpoints and poses are not normalized.}
    \label{fig:app_independent_runs}
\end{figure}
\FloatBarrier

\subsection{Training and Revision Effort}
\label{app:compute}

All five runs acquired wall-supported elevation, supported manipulation, and wall descent within a budget of $1.5\times10^9$ environment steps per controller. B02 and B03 then reused acquired controllers without further training. Table~\ref{tab:app_agent_resources} summarizes the resources used by each run, including unsuccessful pilots, restarts, and resumed segments. Training computation sums logged rollout-collection and learning times, excluding startup, evaluation, task execution, and agent work. It averages $5.5\pm0.9$~h across the five runs.

Across the five Claude runs, training takes 1.2--1.7~h for wall-supported elevation, 2.6--3.5~h for supported manipulation, and 0.4--2.0~h for wall descent. Supported manipulation accounts for the largest share of training time in every completed run.
Controller acquisition involved iterative changes to the training formulation. The five runs required 1--5 controller-design revisions ($3.4\pm1.5$ per run). These counts include changes to rewards, curricula, or termination conditions and exclude documentation changes and implementation or infrastructure repairs.

\begin{table}[!ht]
    \centering
    \caption{Resources per acquisition run. Training includes pilots, restarts, and resumed segments; elapsed time covers A01--B03. Model costs are estimates excluding GPU compute.}
    \label{tab:app_agent_resources}
    \footnotesize
    \setlength{\tabcolsep}{9pt}
    \begin{tabular}{@{}cccccc@{}}
        \toprule
        Run & \shortstack{Training steps ($10^9$)} & \shortstack{Training (h)} & \shortstack{Controller-design revisions} & \shortstack{Elapsed (h)} & \shortstack{Model cost (USD)} \\
        \midrule
        Claude~1 & 1.30 & 5.2 & 3 & 21.9 & \$343 \\
        Claude~2 & 1.31 & 5.5 & 1 & 20.0 & \$258 \\
        Claude~3 & 1.54 & 6.2 & 5 & 21.3 & \$315 \\
        Claude~4 & 0.98 & 4.2 & 4 & 22.2 & \$268 \\
        Claude~5 & 1.55 & 6.3 & 4 & 20.7 & \$253 \\
        \bottomrule
    \end{tabular}
\end{table}

\subsection{Runtime, Agent Usage, and Estimated Cost}
\label{app:agent_usage}

The full A01--B03 sequence took 20.0--22.2~h across the five runs. Elapsed time includes controller acquisition, task-program development, and evaluation.
Estimated model usage costs range from \$253 to \$343 per run, excluding GPU compute.

\FloatBarrier

\subsection{Failure Patterns and Interpretation}
\label{app:run_failures}

\noindent\textbf{Task failures and integration.} Claude Run~5's A01 variant~38 timed out short of a low goal; its B03 variant~3 ended holding the object without entering elevation, with wall heading just outside the switch guard (Appendix~\ref{app:reuse_examples}). Claude Run~3 required an A04 controller hand-back routing repair without additional training. Development repair successes may come from different program versions; the held-out counts evaluate each selected frozen program and library snapshot.

\noindent\textbf{Scope of the evidence.} The study tests acquisition within two supplied training templates and reuse on related tasks under a shared starting configuration. Independent repositories and empty acquired memories establish separation between runs. Development failures motivate capability hypotheses but do not establish that every possible program using the initial library would fail. Controller confirmation reuses development variants; held-out task evaluation measures unseen variants under nominal simulation physics.